\documentclass[10pt,twocolumn,letterpaper]{article}

\usepackage{wacv}
\usepackage{times}
\usepackage{epsfig}
\usepackage{epstopdf}
\usepackage{graphicx}
\usepackage{amsmath}
\usepackage{amssymb}
\usepackage{booktabs} %for better rules in the table
\usepackage{textcomp}
\usepackage{stackengine}
\usepackage{tabularx}

\wacvfinalcopy % *** Uncomment this line for the final submission

\def\wacvPaperID{554} % *** Enter the wacv Paper ID here

\ifwacvfinal\fi
\begin{document}

%%%%%%%%% TITLE
\title{Segmenting Sky Pixels in Images}

% Authors at the same institution
%\author{First Author \hspace{2cm} Second Author \\
% Authors at different institutions

\author{\thanks{First two authors contributed equally.} \ Cecilia La Place\\
Arizona State University\\
{\tt\small cecilia.laplace@asu.edu}
\and
\footnotemark[1]
 Aisha Urooj Khan\\
University of Central Florida\\
{\tt\small aishaurooj@gmail.com}
\and
Ali Borji\\
University of Central Florida\\
{\tt\small aliborji@gmail.com}
}

\maketitle
\ifwacvfinal\thispagestyle{empty}\fi

%%%%%%%%% ABSTRACT
\begin{abstract}
   Outdoor scene parsing models are often trained on ideal datasets and produce quality results. However, this leads to a discrepancy when applied to the real world. The quality of scene parsing, particularly sky classification, decreases in night time images, images involving varying weather conditions, and scene changes due to seasonal weather. This project focuses on approaching these challenges by using a state-of-the-art model in conjunction with non-ideal datasets: SkyFinder and a subset from the SUN database containing the Sky object. We focus specifically on sky segmentation, the task of determining sky and not-sky pixels, and improving upon an existing state-of-the-art model: RefineNet. As a result of our efforts, we have seen an improvement of 10-15\% in the average MCR compared to the prior methods on the SkyFinder dataset. We have also improved from an off-the-shelf model in terms of average mIOU by nearly 35\%. Further, we analyze our trained models on images w.r.t two aspects: times of day and weather, and find that in spite of facing the same challenges as prior methods, our trained models significantly outperform them.
\end{abstract}

%TODO - needs to be referenced in the text
\begin{figure}[t]
\begin{tabular}{l}

\hspace{-8pt} \includegraphics[width=1\linewidth,height=1.5in]{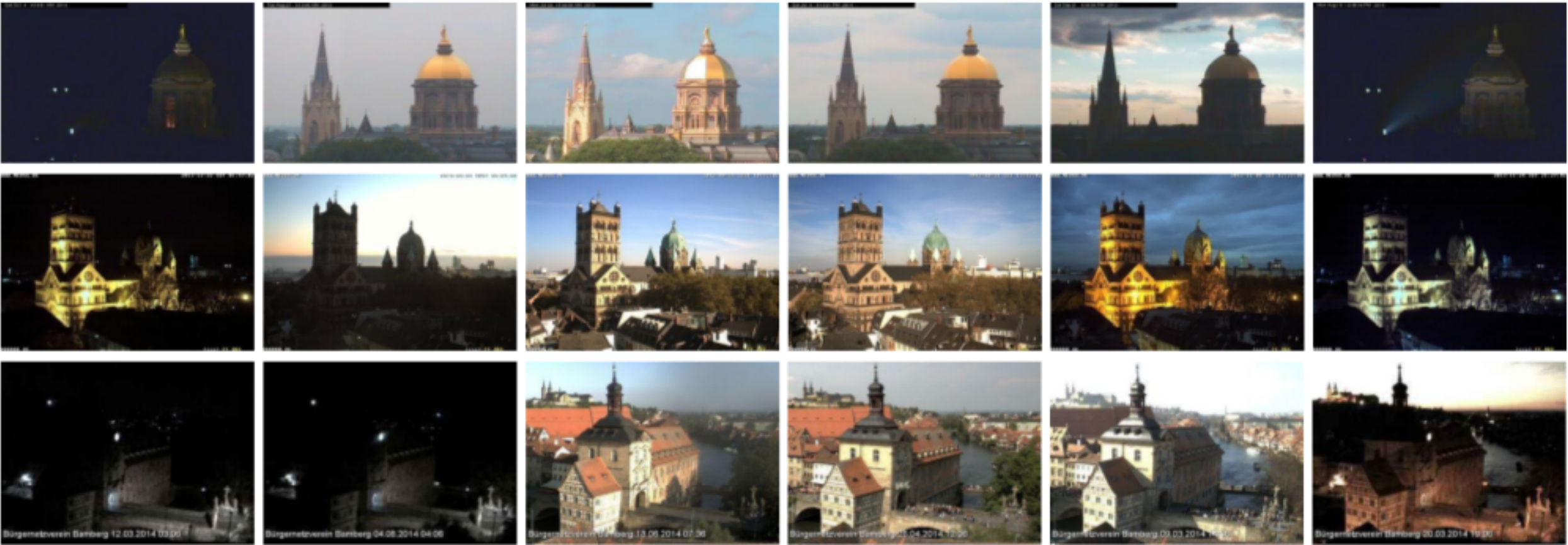} \\
\midrule
\hspace{-8pt} \includegraphics[width=1\linewidth,height=1.5in]{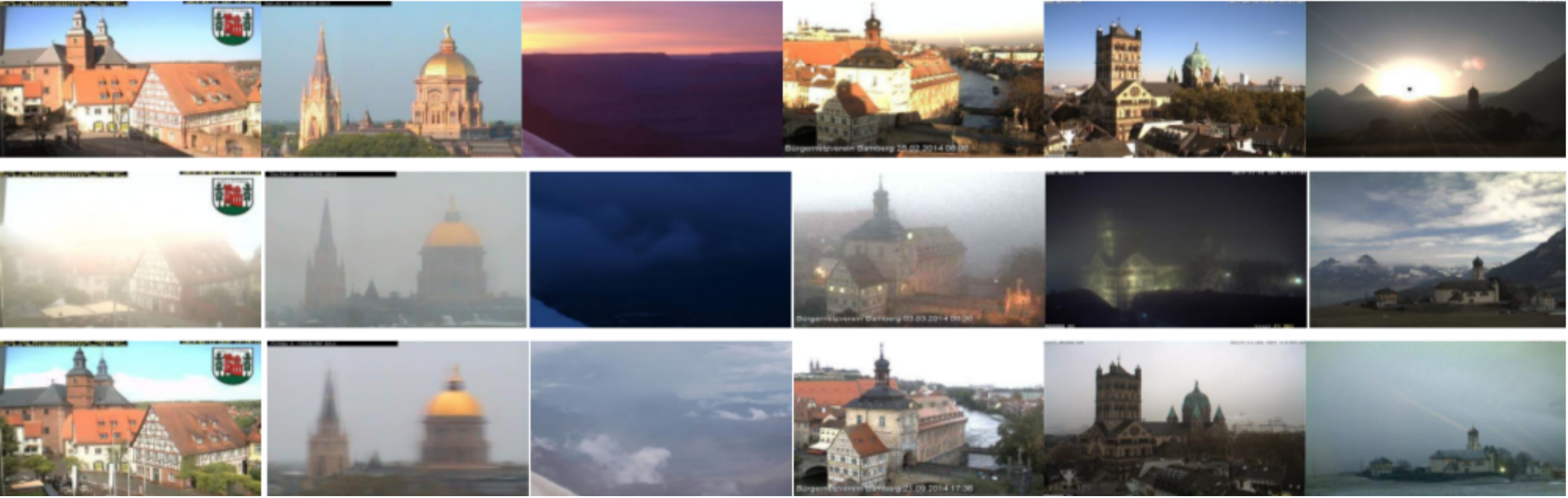} \\
\midrule
\hspace{-8pt} \includegraphics[width=1\linewidth,height=0.5in]{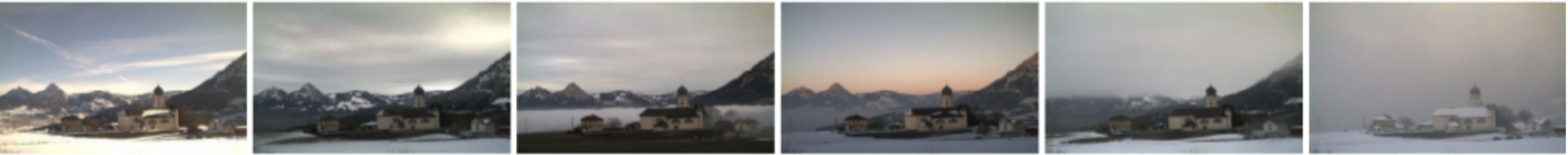} \\
\end{tabular}

\caption{\small \textbf{SkyFinder dataset}. Top 3 rows show sample images for different times of day. Each column signifies a different time: 12 am, 4am, 8am, 12pm, 4pm and 8pm, across 3 separate cameras. Rows 4, 5 and 6 show images across different scenes for different weather types: clear, cloudy, and fog respectively, whereas, the last row shows different weather conditions (clear, cloudy, hazy, rain and sleet) over the same scene.}
\label{fig:timeCompiled}
\vspace{-15pt}
\end{figure}

% %TODO - needs to be referenced in the text
% \begin{figure}[h]
% `\includegraphics[width=1\linewidth,height=1.5in]{images/compiledWeather.png}
% \caption{Each row signifies a different weather type: clear, rain, and partly cloudy, across 3 separate cameras.}
% \label{fig:weatherCompiled}
% \end{figure}

%%%%%%%%% BODY TEXT
\section{Introduction}

Sky segmentation is a part of the scene parsing world in which algorithms seek to label or identify objects in images. However, due to being trained on ideal datasets, these algorithms face difficulty in non-ideal conditions \cite{C1}. As deep learning methods might become more involved in real world applications, it becomes apparent that off-the-shelf methods are not always effective and reliability comes into question. 
Inspired by Mihail et al., \etal\cite{C1} who compared existing sky segmentation methods and sought to bring attention to the problem of ideal datasets, we focused upon approaching the challenges they mentioned through existing models. The challenges of outdoor scene parsing lie in the time of day, year, and varying weather conditions. Their SkyFinder dataset allowed us to pursue these challenges in order to obtain improved results. 
This work highlights the importance of this problem. We offer an improved model that will aid in challenging sky segmentation in the real world. 
In this work, we adopted state-of-the-art segmentation model \cite{C11} for the task of pixel level detection of the sky. Our task is different than semantic segmentation in a sense that we are only interested in one object i.e. sky. Sky, unlike other objects can be difficult to segment due to poor lightning (night time) and weather conditions where even humans are likely to fail (e.g., over dense fog, thunder storms, etc). Thus, we attempt to address this problem in this work.

Our contribution is as following. \textbf{First}, we evaluated an off-the-shelf state-of-the-art model \cite{C11} to demonstrate that existing models fail for different weather conditions, times of day, and other transient attributes \cite{C1}. \textbf{Second}, we fine-tuned RefineNet-Res101-Cityscapes model on the SkyFinder dataset and obtained 12.26\% improvement over off-the-shelf model in terms of misclassification rate (MCR). \textbf{Third}, taking advantage of the existing big dataset, we trained \cite{C11} solely on SkyFinder, which further improved accuracy and outperformed all baseline methods. \textbf{Fourth}, to study across datasets performance of our trained models, we selected a subset from SUN~\cite{xiao2010sun} dataset with the 'sky' label. We then both fine-tuned and trained RefineNet-Res101 model on this subset of SUN dataset (we refer to it as the SUN-Sky dataset in this paper), and perform evaluation across models trained on both SkyFinder and SUN-Sky datasets, and report the results. \textbf{Fifth}, following \cite{C1}, we investigate the effect of times of day and weather conditions on performance of our model and transient attributes. \textbf{Sixth}, we compare our analysis with \cite{C1} in terms of MCR and report impact of weather and times of day w.r.t mIOU scores. \textbf{Finally}, we determine the impact of noisy images like motion blur, Gaussian noise, etcetera on our model's performance and report our results on robustness of our approach.

The rest of this paper is organized as follows. In section \ref{related_work}, we present a brief overview of existing works in this area. In section \ref{approach}, we describe details of our plan including utilized datasets, and our models. Section \ref{evaluation} discusses the performance metrics we used to evaluate the trained models. Section \ref{results} presents experimental findings on Sky segmentation task for both datasets and analysis of different attributes, times of day, weather conditions and noise on this task, followed by discussions in section \ref{conclusion}.

\begin{figure}[t]
\centering
\includegraphics[width=1\linewidth,height = 1.8in]{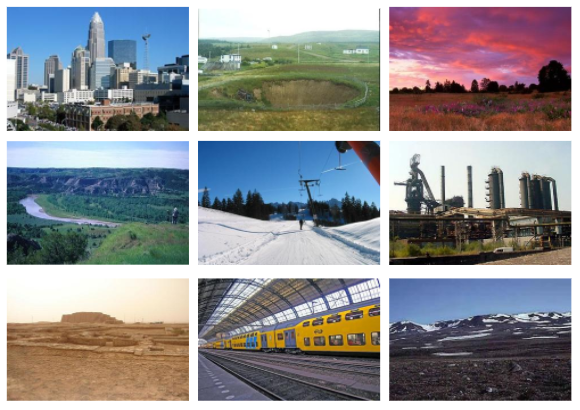}
\label{fig:SUNdb_samples}
\caption{\textbf{SUN-Sky dataset}. Sample images from the various locations described in the SUN dataset with sky.}
\end{figure}
%-------------------------------------------------------------------------

%%%%%%%%%%%%%%%%%% Related work
\section{Related Work} \label{related_work}

The history of scene parsing i.e., assigning each pixel of input image to one of the object labels~\cite{C1}, has evolved from its beginnings. Scene labeling methods \cite{hoiem2005geometric}, \cite{liu2009nonparametric}, \cite{tighe2010superparsing}, \cite{tighe2013finding}, \cite{tighe2014scene} mainly use local appearance information for objects being learned from training data \cite{C1}. Although, previously this task has been addressed by using hand engineered features with a classifier, recent methods address this task by learning features using deep neural networks. From convolutional networks spawned fully convolutional networks which moved away from pixel level algorithms to whole-image-at-a-time methods \cite{C9}. Afterward, the introduction of skip layers led to deep residual learning \cite{C10}. Recently, residual nets became the backbone to scene parsing algorithms such as RefineNet \cite{C11}, PSPNet \cite{C12}, and more. 

Specifically, the sky segmentation task can be helpful for a diverse variety of applications such as stylizing images using sky replacement \cite{tsai2016sky}, obstacle avoidance \cite{de2011sky}, and \cite{roser2008classification} since sky tells a lot about weather conditions. Current applications of sky segmentation range from personal to public use and more. Often times they are used in scene parsing \cite{C2, C3}, horizon estimation, and geolocation. Other applications include weather classification \cite{C4}, image editing \cite{C5,tao2009skyfinder}, weather estimation \cite{C7, C8}, and more. \cite{chu2017camera} and \cite{yan2009weather} worked on weather recognition and used camera as weather sensors. Weather detection has also been used for image searching \cite{tao2009skyfinder} where one can search outdoor scene images based on weather related attributes.

Dev \etal used color based methods \cite{dev2017color} for segmentation of sky/cloud images, whereas \cite{yazdanpanah2013sky} proposed deep learning approach for segmenting sky. Like \cite{C1}, we used the same baseline methods (Hoeim \etal \cite{C2}, Tighe \etal \cite{C3} and Lu \etal \cite{C4}) to compare our results. Hoiem \etal uses a single image to produce an estimate of scene geometry for three classes: ground, sky and vertical regions by learning underlying geometry of an image via appearance-based models. Tighe \etal combines region-level features with SVM-based sliding window detectors for parsing an image into different class labels including 'sky'. Lu \etal classifies an input image into two classes: sunny or cloudy. Their work uses sky detection as an important cue for weather classification. For detecting sky, they used random forests to produce seed patches for sky and non-sky, and then used graph cut to segment sky regions. \cite{C1} uses their sky detector for reporting results which we also used for our comparisons on the SkyFinder dataset. Here we investigate the effectiveness of existing state-of-the-art segmentation methods for this specific problem. To select among the best contenders for this task, we evaluated off-the-shelf RefineNet \cite{C11} and PSPNet \cite{C12} methods on SkyFinder dataset, and chose \cite{C11} for our further experiments as it outperforms \cite{C12} with a large margin on SkyFinder dataset. We further take a deep insight on challenges (such as weather conditions, night time images, and noisy images) which are faced even when robust methods are used.

% \begin{table*}[ht]
% \begin{center}
% \renewcommand{\tabcolsep}{1mm}
% \begin{tabular}{|l|c|c|c|c|c|c|c|c|}
% \hline
% Method & S1-MCR & S1-mIOU & S2-MCR & S2-mIOU & S3-MCR & S3-mIOU & ave-MCR & ave-mIOU\\
% \hline\hline
% Hoiem \etal & 21.28\% & - & 20.68\% & - & 26.24\% & - & 22.73\% & -\\
% Lu \etal & 25.38\% & - & 21.67\% & - & 23.32\% & - & 20.38\% & -\\
% Tighe \etal & 17.48\% & - & 20.33\% & - & 31.58\% & - & 26.21\% & -\\
% RefineNet-Res101-Cityscapes & 17.12\% & 49.31\% & 18\% & 46.31\% & 21.68\% & 49.87\% & 18.93\% & 48.5\%\\
% RefineNet-Res101-SkyFinder-FT & 5.17\% & 79.48\% & 7.48\% & 72.18\% & 7.37\% & 85.2\% & 6.67\% & 78.95\%\\
% RefineNet-Res101-SkyFinder & \textbf{5.08\%} & \textbf{87.07}\%\ & \textbf{7\%} & \textbf{73.84\%} & \textbf{5.65\%} & \textbf{88.05\%} & \textbf{5.9\%} & \textbf{82.99\%}\\
% \hline
% \end{tabular}
% \end{center}

% \caption{\small Results from all three testing splits. MCR results for top 3 baselines are used from the SkyFinder metadata in order to compare our pixel-level sky detector with their methods. We also report mIOU scores. RefineNet-Res101-Cityscapes is the off-the-shelf model trained on Cityscapes. RefineNet-Res101-SkyFinder-FT shows results when we fine-tuned RefineNet-Res101-Cityscapes model on SkyFinder dataset. Finally, training RefineNet-Res101 from scratch on SkyFinder dataset (last row) gives best results. For RefineNet, all these numbers are reported when the model was evaluated at test scale of 0.8.}
% \label{tab:sky-segment-results}
% \end{table*}

\begin{table*}[t]
\small
  \centering
  \setlength\tabcolsep{2pt}
  \renewcommand{\arraystretch}{.9}
  
    \begin{tabular*}{\textwidth}{l @{\extracolsep{\fill}} cccccccc}
        \toprule
                \multicolumn{1}{c}{\textbf{Method}} & \multicolumn{2}{c}{\textbf{Split 1}} & \multicolumn{2}{c}{\textbf{Split 2}} & \multicolumn{2}{c}{\textbf{Split 3}} & \multicolumn{2}{c}{\textbf{Avg.}}\\
                
                \cmidrule(lr){2-3} \cmidrule(lr){4-5} \cmidrule(lr){6-7} \cmidrule(lr){8-9}
           &    mIOU(\%) & MCR(\%) & mIOU(\%) & MCR(\%) & mIOU(\%) & MCR(\%) & mIOU(\%) & MCR(\%)\\    
        
        \midrule
        \textbf{Hoiem \etal} &  - & 21.28 & - & 20.68 & - & 26.24  & - &  22.73\\
        \textbf{Lu \etal} &  - & 25.38 &  - & 21.67 &  - & 23.32  & - & 20.38\\
        \textbf{Tighe \etal} &  - & 17.48 & - & 20.33 & - & 31.58 & - &26.21\\
        \textbf{RefineNet-Res101-Cityscapes} &  49.31 & 17.12 & 46.31 &18.00 &  49.87 & 21.68 & 48.5&18.93 \\
        \textbf{RefineNet-Res101-SkyFinder-FT } & 79.48 &5.17 &   72.18 &7.48 & 85.2 &  7.37 & 78.95& 6.67 \\
        \textbf{RefineNet-Res101-SkyFinder} &  \textbf{87.07}\ & \textbf{5.08} & \textbf{73.84} &\textbf{7.00} & \textbf{88.05} &  \textbf{5.65} & \textbf{82.99}& \textbf{5.90} \\

        \bottomrule
    \end{tabular*}
    \vspace*{5pt}
   \caption{\small Results from all three testing splits. MCR results for top 3 baselines are used from the SkyFinder metadata in order to compare our pixel-level sky detector with their methods. We also report mIOU scores. RefineNet-Res101-Cityscapes is the off-the-shelf model trained on Cityscapes. RefineNet-Res101-SkyFinder-FT shows results when we fine-tuned RefineNet-Res101-Cityscapes model on SkyFinder dataset. Finally, training RefineNet-Res101 from scratch on SkyFinder dataset (last row) gives best results. For RefineNet, all these numbers are reported when the model was evaluated at test scale of 0.8.}
\label{tab:sky-segment-results}
      %  \vspace{0pt}
\end{table*}

\section{Approach} \label{approach}

Our approach is based on adopting semantic segmentation methods for the task of pixel-level sky detection. We used two datasets: SkyFinder \cite{C1} and SUN-Sky \cite{xiao2010sun} for our work. Our approach outperforms baseline methods in the task of sky segmentation. We evaluated generalization capability of our models by performing across datasets evaluation. Also, we studied influence of various factors like transient attributes, weather conditions and noise on our model's performance. 
\subsection{Datasets} \label{dataset}

\subsubsection{SkyFinder dataset}

The SkyFinder dataset is a subset of the Archive of Many Outdoor Scenes (AMOS) dataset. Due to the full SkyFinder dataset not being available, we used 45 of the 53 cameras shared. This entailed ~60K-70K images, with an average of ~1500 images per camera. These images were of varying dimensions, quality, time of day, season, and weather. However, some cameras contained images which indicated the camera was experiencing technical difficulties or repairs. These few images were removed to focus on the challenges we wished to address. We then changed the sizes of the images to introduce a form of uniformity and make test evaluation faster by resizing them to be within the ranges of $320 \times 480$. 

A single segmentation map was associated with each camera. This is due to each camera being stationary for at least a year in order to be included within in the dataset. See figure \ref{fig:timeCompiled} for sample images from this dataset.  %We then duplicated the ground truth for each original image, and changed the names of the copies to match the file name of each original image in order to use RefineNet.

\begin{figure}[t]
\begin{center}
\begin{tabular}{ccccc} 

\small Orig. & \small Gt. &  \small a & \small b & \small c\\
\includegraphics[width=0.5in, height=0.38in]{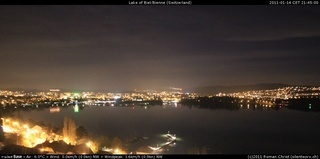} &
\includegraphics[width=0.5in, height=0.38in]{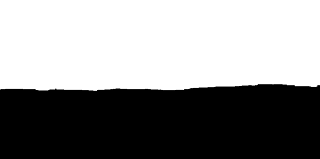} &
\includegraphics[width=0.5in, height=0.38in]{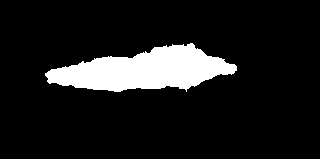} & \includegraphics[width=0.5in, height=0.38in]{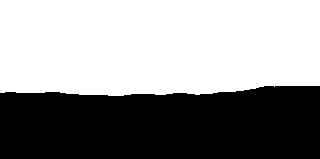} & \includegraphics[width=0.5in, height=0.38in]{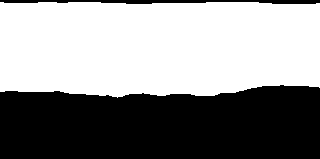}\\
\includegraphics[width=0.5in]{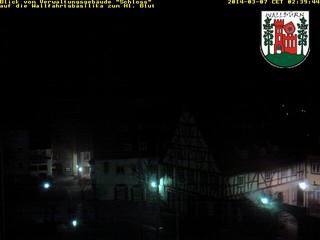} &
\includegraphics[width=0.5in]{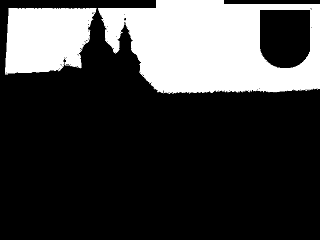} &
\includegraphics[width=0.5in]{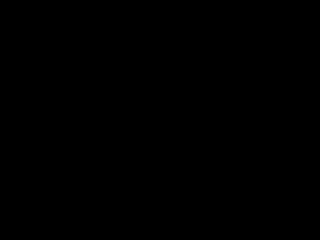} & \includegraphics[width=0.5in]{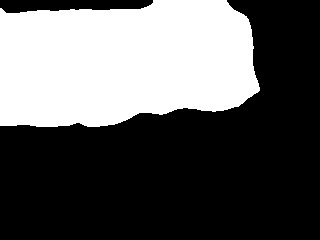} & \includegraphics[width=0.5in]{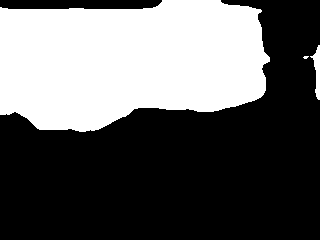}\\
\includegraphics[width=0.5in]{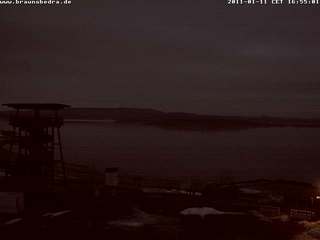} &
\includegraphics[width=0.5in]{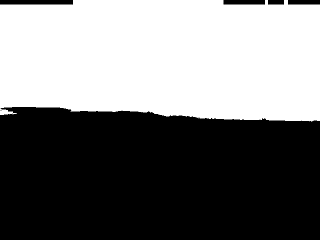} &
\includegraphics[width=0.5in]{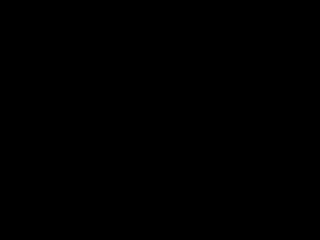} & \includegraphics[width=0.5in]{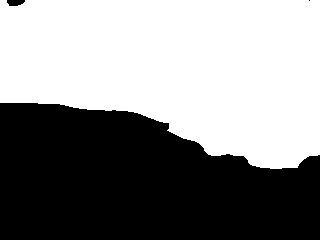} & \includegraphics[width=0.5in]{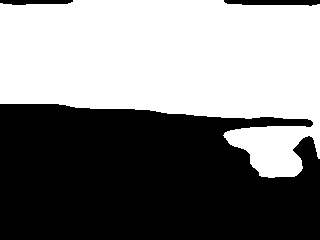}\\
\end{tabular}
\end{center}
\caption{\small \textbf{Improvement in night time images}. Column 1 shows original image examples, col2 shows ground truths, where last three columns show results for a) off-the-shelf RefineNet-Res101-Cityscapes, b) RefineNet-Res101-SkyFinder-FT, and c) RefineNet-Res101-SkyFinder respectively.  }
\label{fig:night-time-improvement}
\vspace{0pt}
\end{figure}

\begin{figure}[t]
\begin{center}
\begin{tabular}{ccccc}
\hspace{10pt} \small Orig. & \hspace{20pt} \small Gt. & \hspace{25pt} \small a & \hspace{25pt} \small b & \hspace{10pt} \small c\\
\multicolumn{5}{c}{\includegraphics[width=1\linewidth,height=1.5in]{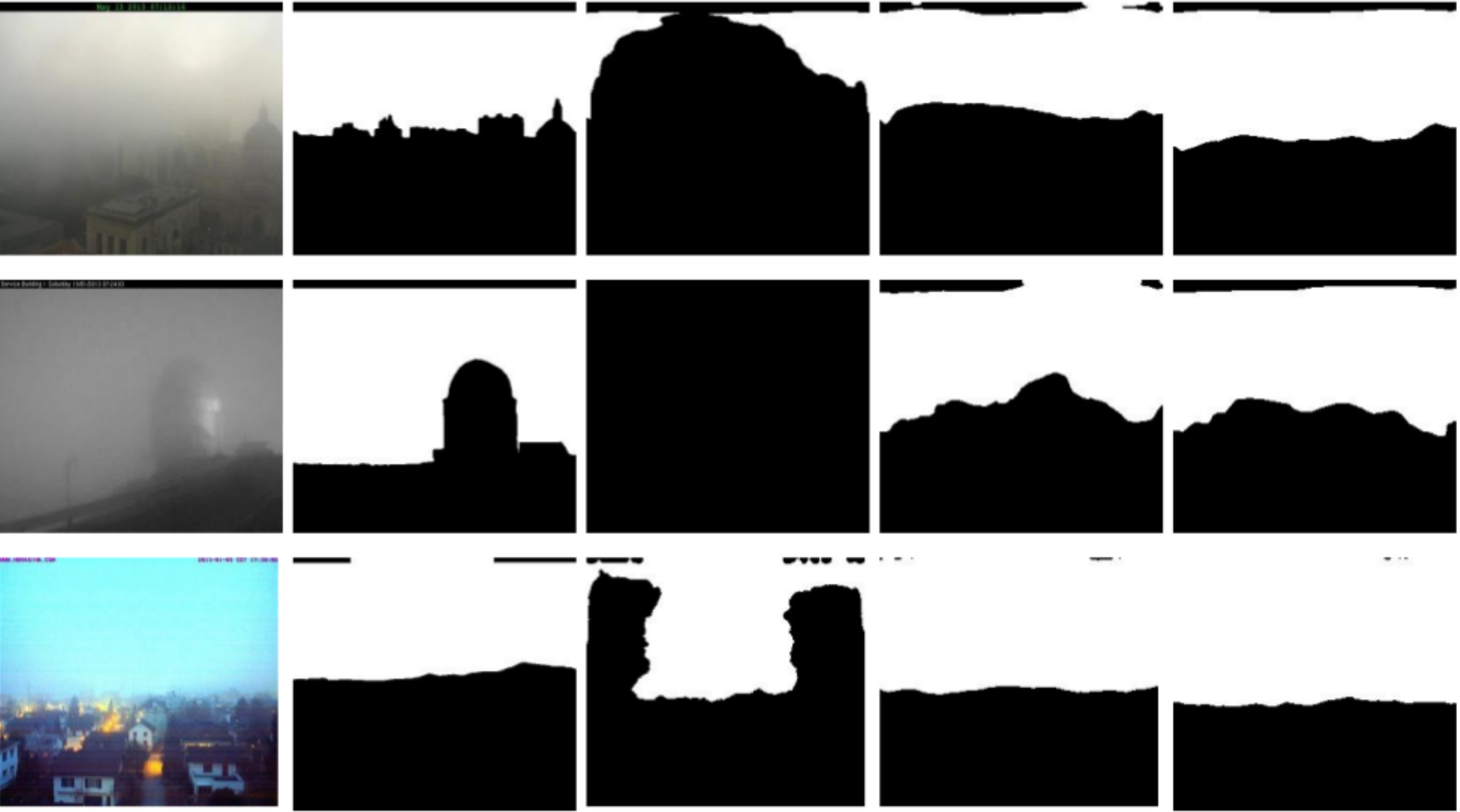}}

\end{tabular}
\end{center}
\caption{\small \textbf{Improvement in weather obscured images}. Column 1 shows original image examples, col2 shows ground truths, where last three columns show results for a) off-the-shelf RefineNet-Res101-Cityscapes, b) RefineNet-Res101-SkyFinder-FT, and c) RefineNet-Res101-SkyFinder respectively. }
\label{fig:weath-improvement}
\vspace{0pt}
\end{figure}

\subsubsection{SUN dataset} 
The Scene UNderstanding dataset is comprised of a multitude of different scenes and the objects that make up said scenes \cite{C13}. It is constantly being updated through community effort, thus adding to it's already large nature. We primarily looked at this dataset for its "sky" object classification which allowed us to begin comparing the SkyFinder dataset and results. Unlike SkyFinder, images are not grouped by camera, instead they are grouped by scene such as airport terminal, or church. % and furthermore sorted by letter
Also, SkyFinder focused on images taken from stationary cameras, whereas SUN has a variety of images from a variety of viewpoints. For the purposes of this research, we focused on the subsection of the SUN database that had the object "sky" labeled (about 20,000 images).
We resized those images to be within the range of $320 \times 480$ for improved test evaluation speed. In what follows, we refer to this subset as the \textbf{SUN-Sky} dataset. See figure \ref{fig:SUNdb_samples} for few sample images from this dataset. %Table \ref{tab:SUNData} shows quantitative results for SUN-Sky dataset. See fig. \ref{fig:SUNdb_samples} and \ref{fig:SUNdb_segment} for sample images and qualitative results on this dataset. 
% \subsubsection{SUN dataset}

% The SUN dataset consists of a large collection of various types of scenes. While the dataset's original use is for scene recognition \cite{C13}, it entails the classification of objects which allows us to use this dataset scene segmentation. As our work focuses on sky segmentation and the dataset includes classification for sky, we were able to incorporate this dataset into our research. We downloaded the most recent entirety of the dataset in order to isolate only images that contained the sky label and use that in for evaluating cross dataset performance of RefineNet-Res101-SkyFinder.

% After selecting the subset database, we also resized those images to be within the range of 320x480 for improved test evaluation speed. For the rest of this document, we refer to that subset as SUN-Sky dataset. Table \ref{} shows quantitative results for SUN-Sky dataset. See fig. \ref{} for sample images and qualitative results on this dataset.
%------------
\begin{table}
\begin{center}
	\begin{tabular}{l @{\extracolsep{\fill}} cc}
    \toprule
		\textbf{Method}    & \textbf{mIOU(\%)}  & \textbf{MCR(\%)} \\
        \midrule
        RefineNet-Res101-Cityscapes & 61.69 & 8.4\\
        RefineNet-Res101-SUNdb-FT & \textbf{83.1}& \textbf{3.7}\\
        RefineNet-Res101-SUNdb & 82.36 & 4.17 \\
        \bottomrule       
    \end{tabular}
    \end{center}
    \label{tab:SUNData}
    \caption{Performance results of SUNdb Finetune and SUNdb ImageNet.}\label{tab:sundb}
    \vspace{-5pt}
\end{table}
%---------------
\subsection{Sky Segmentation} \label{segment}

\subsubsection{RefineNet}

The model we used, RefineNet, was created by Lin \etal\cite{C11}. This model seeks to to retain detail throughout the reconstruction of the image and its segmented output unlike its predecessor and backbone, ResNet. 

\subsubsection{Off-the-Shelf RefineNet}

In order to establish a baseline in RefineNet, we used RefineNet's Res101 model that was trained on Cityscapes, a dataset of European cities for scene segmentation. The Cityscapes dataset is an ideal dataset for the Sky class, and as a result does well in ideal conditions in the SkyFinder dataset. After setting up the model and using it on a single Titan X GPU, we ran each of the 45 cameras through the model and evaluated solely on their sky classification ability, i.e. all other classes were construed as non-sky. We refer this baseline model as RefineNet-Res101-Cityscapes in the following.

\subsubsection{Finetuning}

To obtain proof of concept prior to running the entire dataset we focused on a smaller subset of the dataset. We took 10 random cameras and broke it into a train-val-test split. From each camera we took 75 images for training, 25 images for validation, and between 175-300 images for testing and evaluation.

Following the success of fine-tuning the model on the subset, we fine-tuned the RefineNet-Res101-Cityscapes model on the SkyFinder dataset. Being unable to find the same train-val-test split as Mihail \etal \cite{C1}, we split the dataset into our own train-val-test splits.
%we fine-tuned on the model trained on Cityscapes by initializing from it and using RefineNet's code along with a different train-val-test split than our proof of concept version and the one incompletely referred to in Mihail \etal\cite{C1} as it was not clarified which were used.
 Although, to keep our experiments as much consistent as possible to \cite{C1}, we used the same number of test cameras in our experiments. Hence, our split consisted of 13 cameras used for testing, 4 cameras used for validation, and the remaining cameras used for training. We then shuffled the cameras in each section but kept the same number of cameras for each training, validation, and testing set, and repeated the aforementioned fine-tuning process two more times for a total of three fine-tuning trials. We used a learning rate of 5e-5 for each instance and trained the model for 10 epochs. After 10 epochs, the validation accuracy started leveling out. Thus, for consistency, we report all results on model trained till 10 epochs. We refer our fine-tuned models as RefineNet-Res101-SkyFinder-FT in our results.

\subsubsection{Training with SkyFinder dataset}

Finally, to take advantage of the big size of the dataset, we trained RefineNet-Res101 from scratch, where Res101 was initialized with pre-trained ImageNet model but trained the RefineNet on solely the SkyFinder dataset. We used the same three train-val-test splits as mentioned above (to allow for fair comparison) and trained at a learning rate of 5e-4 for over 10 epochs. Due to the flattening of the learning curve, we used the model at epoch 10 for testing in both instances. We refer to these models as RefineNet-Res101-SkyFinder in our results.

\subsubsection{Training with Sun-Sky dataset}

We broke down the sky-labeled sub-dataset of the SUN dataset into a randomly shuffled 60-20-20 split for our uses. After resizing the images to $320 \times 240$ to train and test quickly, we then generated the ground truth segmentation masks by keeping only the sky class and treating any other classes as non-sky. To train and evaluate, we focused on a similar process as we did with the SkyFinder dataset. Training entails fine-tuning from the RefineNet-Res101-Cityscapes model, and training on a model initialized from ImageNet-Res101. Evaluation is comprised of calculating the average MCR and mIOU of the dataset. We finetuned RefineNet-Res101-Cityscapes model using the SUN-Sky dataset for 10 epochs at a learning rate of 5e-4, and subsequently 10 more epochs at a learning rate of 5e-5. For our second model, we initialized from ImageNet-Res101 and trained using the same split as we did for the previous model for 10 epochs at a learning rate of 5e-4. Much like the previous model, we again trained another 10 epochs at a lower learning rate of 5e-5. 

\begin{figure}[t]
\centering
\begin{tabular}{ccccc}
\hspace{10pt} \small Orig. & \hspace{20pt} \small Gt. & \hspace{25pt} \small a & \hspace{30pt} \small b & \hspace{10pt} \small c\\
\multicolumn{5}{c}{\includegraphics[width=1\linewidth,height = 1.8in]{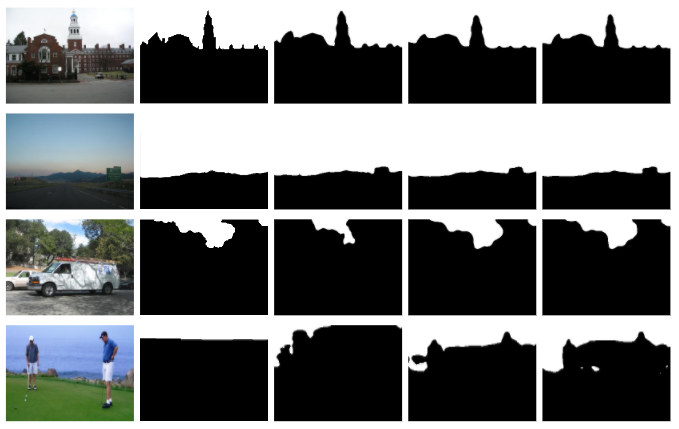}}\\
\end{tabular}
\label{fig:SUNdb_segment}
\caption{Segmentation improvement across the RefineNet-Res101-Cityscapes model (a), the RefineNet-Res101-SUN-FT model (b), and the RefineNet-Res101-SUN model (c).}
\end{figure}

Table \ref{tab:sundb} shows quantitative results for SUN-Sky dataset on both fine-tuned model (RefineNet-Res101-SUNdb-FT) and trained model (RefineNet-Res101-SUNdb). See fig. \ref{fig:SUNdb_segment} for qualitative results on this dataset. 

%-----------------------------------------------------------

\begin{table}[t]
\scriptsize
  \centering
  \setlength\tabcolsep{.1pt}
  \renewcommand{\arraystretch}{.9}
  
    \begin{tabular*}{.48\textwidth}{c @{\extracolsep{\fill}} cccccccc}
        \toprule
                \multicolumn{1}{c}{\textbf{Time of day}} & \multicolumn{2}{c}{\textbf{Split 1}} & \multicolumn{2}{c}{\textbf{Split 2}} & \multicolumn{2}{c}{\textbf{Split 3}} & \multicolumn{2}{c}{\textbf{Avg.}}\\
                
                \cmidrule(lr){2-3} \cmidrule(lr){4-5} \cmidrule(lr){6-7} \cmidrule(lr){8-9}
       \textbf{(hour)}    &    mIOU & MCR & mIOU & MCR & mIOU & MCR & mIOU & MCR\\   
           
        \midrule
    \textbf{0} & 82.36 & 5.54 & 80.64 & 6.27 & 85.37 & 7.46 & 82.79 & 6.42\\
    \textbf{1} & 76.28 & 8.25 & 75.40 & 7.69 & 84.88 & 7.34 & 78.85 & 7.76\\
    \textbf{2} & 76.12 & 9.75 & 76.26 & 8.31 & 86.27 & 6.75 & 79.55 & 8.27\\
    \textbf{3} & 70.41 & 7.21 & 69.99 & 6.28 & 87.96 & 5.97 & 76.12 & 6.48\\
    \textbf{4} & 71.66 & 6.65 & 69.83 & 5.71 & 88.26 & 5.06 & 76.58 & 5.81\\
    \textbf{5} & 77.33 & 5.16 & 71.04 & 4.84 & 90.82 & 3.09 & 79.73 & 4.36\\
    \textbf{6} & 76.48 & 5.06 & 74.20 & 5.85 & 91.74 & 3.94 & 80.80 & 4.95\\
    \textbf{7} & 76.66 & 3.65 & 71.05 & 8.71 & 89.62 & 5.75 & 79.11 & 6.04\\
    \textbf{8} & 72.05 & 3.31 & 69.53 & 4.15 & 93.23 & 3.46 & 78.27 & 3.64\\
    \textbf{9} & 72.47 & 3.06 & 70.65 & 3.34 & 93.93 & 2.94 & 79.02 & 3.12\\
    \textbf{10} & 74.54 & 2.44 & 73.21 & 2.89 & 94.51 & 2.69 & 80.75 & 2.68 \\
    \textbf{11} & 70.97 & 2.57 & 68.90 & 2.82 & 94.17 & 2.89 & 78.01 & 2.76\\
    \textbf{12} & 74.08 & 2.61 & 72.69 & 2.92 & 94.57 & 2.57 & 80.44 & 2.70\\
    \textbf{13} & 74.04 & 2.12 & 71.78 & 2.61 & 94.60 & 2.49 & 80.14 &2.41 \\
    \textbf{14} & 72.58 & 2.41 & 70.38 & 2.70 & 94.17 & 2.81 & 79.04 & 2.64\\
    \textbf{15} & 73.25 & 2.40 & 71.08 & 2.85 & 94.19 & 2.84 & 79.50 & 2.70\\
    \textbf{16} & 71.69 & 2.59 & 69.78 & 2.86 & 92.77 & 3.44 & 78.08 & 2.96\\
    \textbf{17} & 71.73 & 2.85 & 69.63 & 3.13 & 91.76 & 3.73 &  77.71&3.24 \\
    \textbf{18} & 70.12 & 3.33 & 67.50 & 3.51 & 90.41 & 4.37 & 76.01 & 3.73\\
    \textbf{19} & 69.00 & 3.96 & 66.53 & 4.42 & 89.38 & 5.00 & 74.97 & 4.46\\
    \textbf{20} & 69.43 & 5.74 & 64.92 & 11.19 & 85.27 & 8.08 & 73.21 & 8.34\\
    \textbf{21} & 67.11 & 5.70 & 64.85 & 6.79 & 86.80 & 5.88 & 72.92 & 6.12\\
    \textbf{22} & 68.27 & 6.58 & 64.70 & 6.09 & 86.50 & 5.49 & 73.16 & 6.05\\
    \textbf{23} & 70.86 & 7.17 & 62.64 & 6.48 & 83.62 & 6.89 & 72.37 & 6.85\\
      \midrule
     \textbf{Avg.} & 72.89 & 4.77 & 70.30 & 6.45 & 90.20 & 4.62 & 77.8 & 5.25 \\
        \bottomrule
    \end{tabular*}
    \vspace*{5pt}
    \caption{\small Sky segmentation results break down w.r.t different times of the day on SkyFinder dataset. Numbers are in percentage.}
    \label{tab:time-of-day-results}
      %  \vspace{0pt}
\end{table}

%------------------------------------------------------------------------

\section{Evaluation} \label{evaluation}

To evaluate the accuracy of our results we used both the misclassification rate (MCR) defined in Mihail \etal\cite{C1}, and the mean intersection over union (mIOU). The use of both of these metrics allowed for the ability to compare our results to the results found in Mihail \etal\cite{C1}, which focused primarily on MCR results, and determine the overlapping accuracy of the segmentation outputs. 

\begin{equation}
    MCR = \frac{FP + FN}{number \ \ of \ \ pixels}
\end{equation}

\begin{equation}
    mIOU = \sum\frac{TP}{TP + FP + FN}
\end{equation}

\section{Results} \label{results}
In this section, we discuss our experiments for sky segmentation, and studying impact of different conditions like times of day, weather situation, transient attributes and noise on performance of our model. Please note that, although \cite{C1} have done a similar study, we extend their work and also report mIOU in our experiments. 

\subsection{SkyFinder dataset}
Since we evaluated RefineNet on SkyFinder dataset in three settings: off-the-shelf RefineNet, fine-tuned, and trained from the scratch, for reference, we use RefineNet-Res101-Cityscapes for off-the-shelf RefineNet model, initialized from Res101 and trained on Cityscapes. We then further fine-tuned same off-the-shelf model on SkyFinder dataset for all three splits and refer to the model as RefineNet-Res101-SkyFinder-FT. RefineNet when initialized from ImageNet pre-trained Res101 and trained on SkyFinder from the scratch is referred as RefineNet-Res101-SkyFinder. We find that RefineNet-Res101-SkyFinder outperforms the fine-tuned model which is clearly because the former takes an advantage of the large size of this dataset. For comparison with other baseline models (Hoiem \etal, Lu \etal, or Tighe \etal) mentioned in \cite{C1}, we used the MCR scores for all three test splits and also report the average performance for all methods in table \ref{tab:sky-segment-results}. To compute MCR scores for our baselines on our test sets, we fetch the MCR score from metadata provided with SkyFinder dataset after they evaluated these models.

For results in table \ref{tab:sky-segment-results}, we evaluated RefineNet at test scale of 0.8 (default setting), which performs better than being evaluated at full scale (scale = 1.0). Please refer to tables \ref{tab:sky-segment-results} and \ref{tab:time-of-day-results} for comparison. We find that RefineNet-Res101-SkyFinder outperforms all baseline methods both in terms of mIOU and MCR scores.  
For qualitative evaluation, the images are selected from a few of the first test set of cameras and do not include visual results from Hoiem \etal, Lu \etal, or Tighe \etal. While Mihail \etal created their own ensemble using the combination of the three methods, their results were unreported for individual images.

Analysis on time of day and weather has been performed with full scale test evaluation.

%TODO - fix table sizing
% \begin{table}[t]
% \scriptsize
%   \centering
%   \setlength\tabcolsep{.1pt}
%   \renewcommand{\arraystretch}{.9}
  
%     \begin{tabular*}{.48\textwidth}{l @{\extracolsep{\fill}} cccccccc}
%         \toprule
%                 \multicolumn{1}{c}{\textbf{Methods}} & \multicolumn{4}{c}{\textbf{Camera 17244}} & \multicolumn{4}{c}{\textbf{Camera 9112}}\\
%                 & \multicolumn{2}{c}{\textbf{Split 1}} & \multicolumn{2}{c}{\textbf{Split 3}} & \multicolumn{2}{c}{\textbf{Split 2}} & \multicolumn{2}{c}{\textbf{Split 3}}\\
                
%                 \cmidrule(lr){2-3} \cmidrule(lr){4-5} \cmidrule(lr){6-7} \cmidrule(lr){8-9}
%            &    mIOU & MCR & mIOU & MCR & mIOU & MCR & mIOU & MCR\\
%            \midrule
%            RefineNet-Res101-SkyFinder-FT & 76.86 & 5.66 & 80.77 & 5.08 & 85.12 & 6.96 & 88.53 & 5.44\\
%            RefineNet-Res101-SkyFinder & 81.40 & 4.81 & 86.50 & 3.45 & 85.56 & 6.76 & 89.71 & 4.93\\
%     \end{tabular*}
%     \caption{Two examples of overlapping cameras' results compared to their Split 3 counterpart. Each camera has been tested on it's own split's model. Numbers are in percentage.}
%     \label{tab:cam-10917}
% \end{table}

\subsubsection{Performance on Camera 10917} First, some background on SkyFinder's Camera 10917: this specific camera is of a location in which the sky does not peek through anywhere. It depicts a quaint village and trees, but no sky. When SkyFinder was trained on the first two splits it did not contain this camera in its training data. As a result we consistently witnessed abysmal IOU values of 0 which results in performance degradation for test splits 1 and 2 because the model has not previously seen images with no sky. However, the model performs decent in terms of MCR values (below 10\%). We believe that the IOU evaluation method in a case such as this is inefficient. Therefore we pay particular attention to the MCR in regards to this camera. For test split 3, this camera has been used for training our models, thus model performs better on the test set.

% However, looking to Table \ref{tab:cam-10917}, we can trace the possible results of the camera's affects across the first and second split in some cameras. In it, we note the mIOU and MCR of cameras that appear in split 3 and split 1 or 2. Of the total 7 overlapping cameras, 5 of them showed improvement in their mIOU and MCR scores when evaluated on the finetuned model. 6 of the 7 showed improvement when evaluated on the model trained from scratch.Improvement ranged from smaller than 1\% to 7\% with regards to mIOU, and 0.5\% to 4\% with regards to the MCR.

% \subsection{SUN dataset}

\subsubsection{Performance for night time}
Fig \ref{fig:night-time-improvement} shows visual improvement in results for night time images when compared with off the shelf RefineNet-Res101-Cityscapes and RefineNet-Res101-SkyFinder fine-tuned, which makes it obvious why our trained models win over the baseline methods in terms of MCR as well.

\subsubsection{Performance for weather obscured images}
In section \ref{weather}, our results suggests that sky segmentation in night time is more challenging than during day hours. 
But, our final trained model on SkyFinder still improves in terms of mIOU over our established baseline models (RefineNet-Res101-Cityscapes and RefineNet-Res101-SkyFinder-FT). Fig. \ref{fig:weath-improvement} shows that despite of images obscured due to dense fog, RefineNet-Res101-SkyFinder was able to perform reasonably well. 
\begin{figure}[t]
\centering
\begin{tabular}{c}
\includegraphics[width=1\linewidth,height = 1.8in]{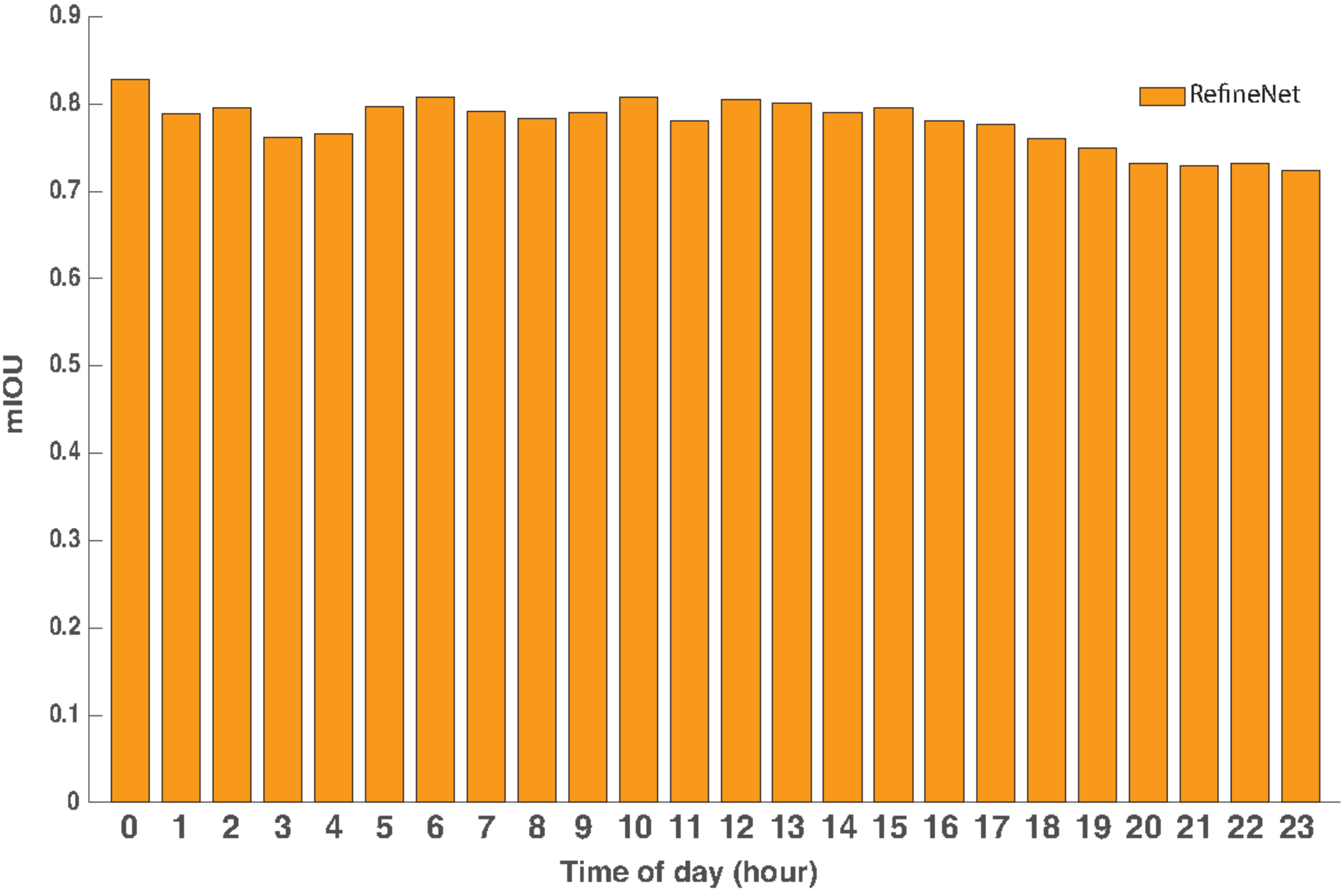}
\\
\includegraphics[width=.9\linewidth,height = 2in]{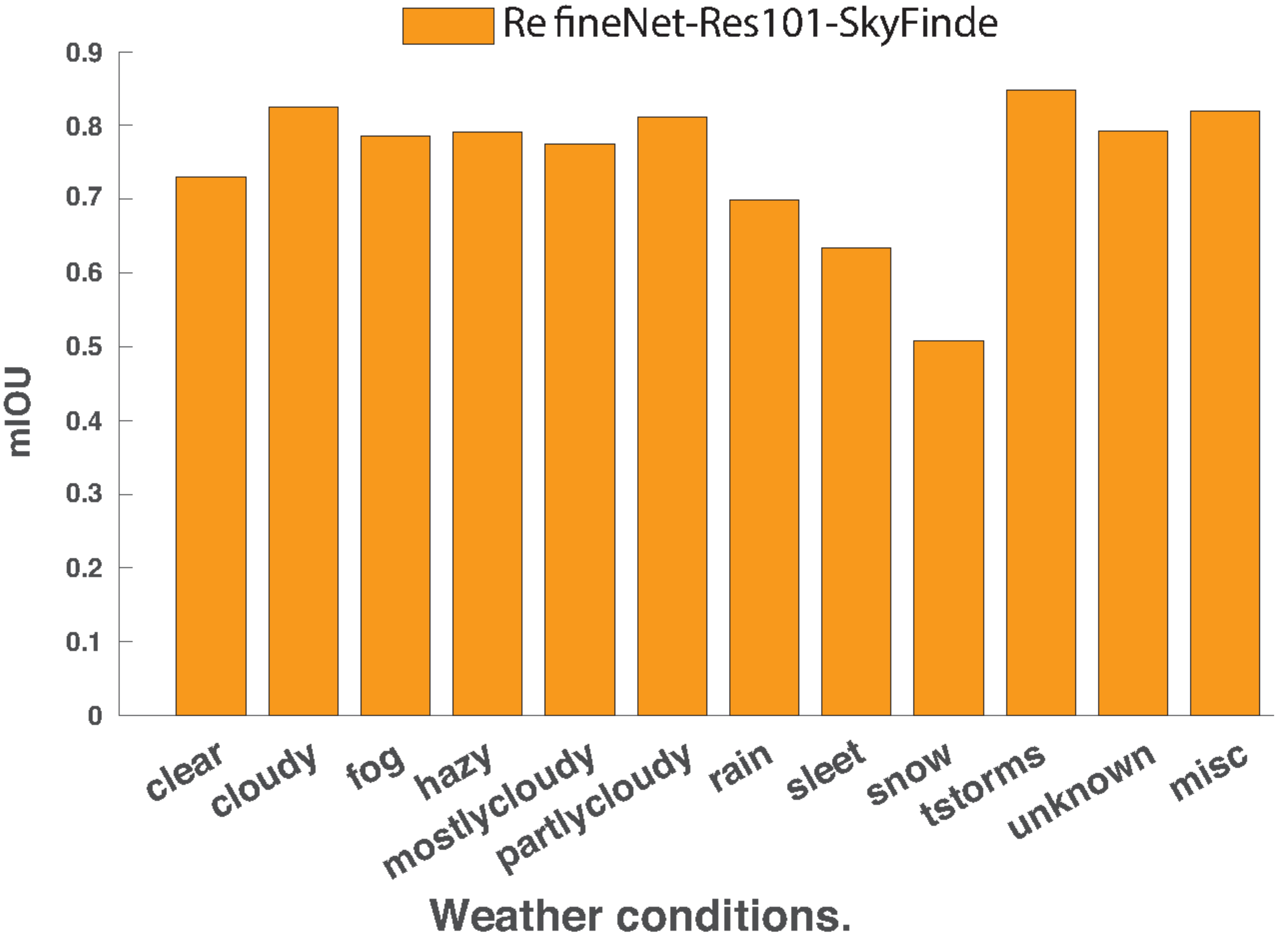}
\end{tabular}
%\vspace*{-10pt}
 \caption{\small Performance analysis of RefineNet-Res101-SkyFinder in terms of mean intersection over union w.r.t row1) time of day, and row2) Weather conditions.  }
\label{fig:iou-weather-tod}
%\vspace{-10pt}
\end{figure}

%------------------------------------------
\begin{figure}[t]
\begin{center}
\includegraphics[width = 1\linewidth,height=2in]{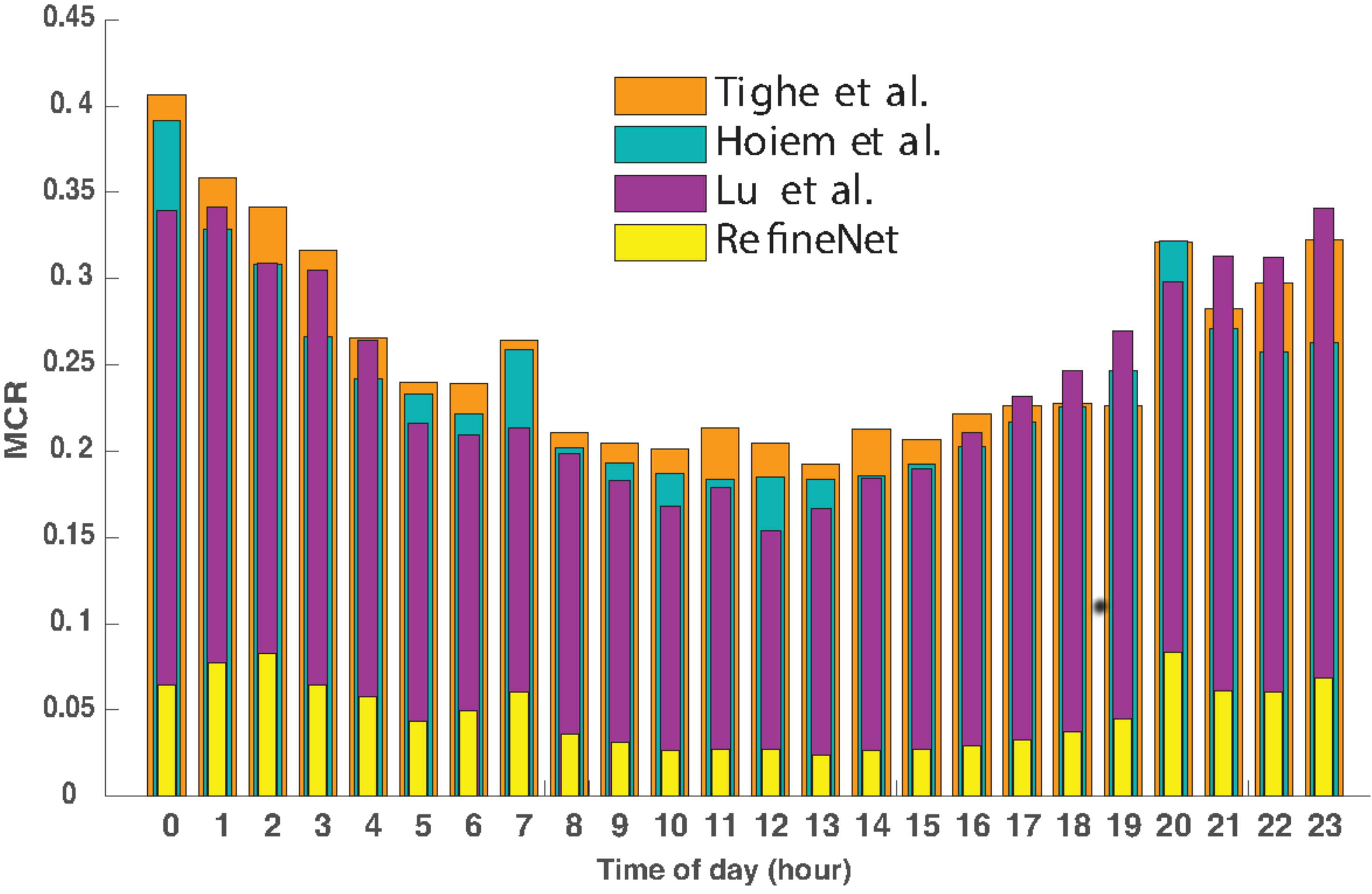}
\end{center}
%\vspace{-15pt}
\caption{\small Illustration for performance analysis on time of day. X-axis shows hour of the day and y-axis represents mean classification rate (MCR) over all three test splits we used in our experiment.}
\label{fig:time-of-day}
%\vspace{-10pt}

\end{figure}

%---------figure for time---
\begin{figure}[t]
\centering
\begin{tabular}{ccccccccc}

\hspace{0pt}Input & \hspace{-5pt} GT & \hspace{-5pt}Pred. &\hspace{-5pt}Input  &\hspace{-5pt} GT & \hspace{-5pt}Pred. & \hspace{-5pt}Input & \hspace{-5pt}GT & \hspace{-5pt}Pred.
\\
\multicolumn{9}{c}{\includegraphics[width=1.0\linewidth]{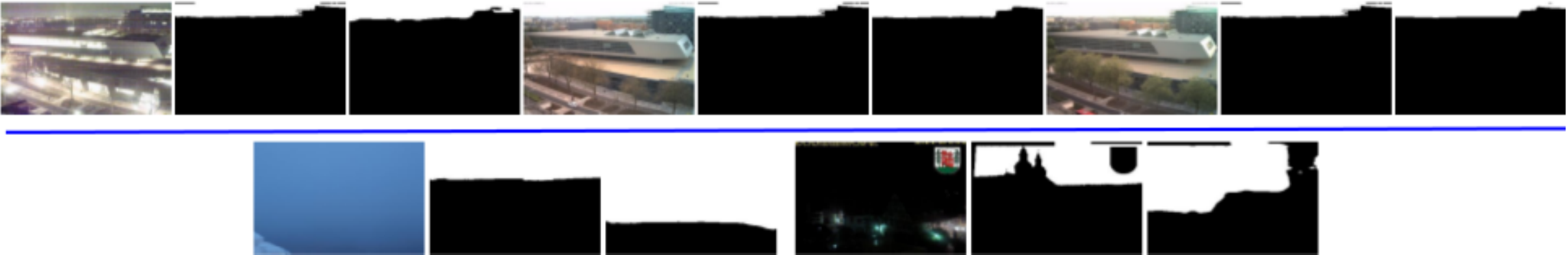}}

\end{tabular}

   \caption{\small Qualitative results for performance analysis w.r.t times of day on SkyFinder dataset using RefineNet-Res101-SkyFinder. The first row shows success cases(mIOU $>$ 0.8) for hour: 0(night), 12(noon) and 18(evening), and the last row shows images where model fails(mIOU $<$ 0.5). The last row shows failure cases for hours 6 and 0. Our model never failed for images around noon.}
\label{fig:time}

\end{figure}
%-------------------------------------------------------------------------

\begin{table}[t]
\scriptsize
  \centering
  \setlength\tabcolsep{.1pt}
  \renewcommand{\arraystretch}{.9}
  
    \begin{tabular*}{.48\textwidth}{l @{\extracolsep{\fill}} cccccccc}
        \toprule
                \multicolumn{1}{c}{\textbf{Weather}} & \multicolumn{2}{c}{\textbf{Split 1}} & \multicolumn{2}{c}{\textbf{Split 2}} & \multicolumn{2}{c}{\textbf{Split 3}} & \multicolumn{2}{c}{\textbf{Avg.}}\\
                
                \cmidrule(lr){2-3} \cmidrule(lr){4-5} \cmidrule(lr){6-7} \cmidrule(lr){8-9}
           &    mIOU & MCR & mIOU & MCR & mIOU & MCR & mIOU & MCR\\    
        
        \midrule
        \textbf{clear} & 59.88 & 2.96 & 66.48 & 3.54 & 92.61 & 4.21 & 72.99 & 3.57\\
        \textbf{cloudy} & 79.92 & 3.61 & 77.02 & 4.19 & 90.71 & 4.07 & 82.55 & 3.96\\
        \textbf{fog} & 77.82 & 7.18 & 71.72 & 7.75 & 86.22 & 5.58 & 78.59 & 6.83\\
        \textbf{hazy} & 75.74 & 6.72 & 72.32 & 6.71 & 89.31 & 4.49 & 79.12 & 5.97\\
        \textbf{mostly cloudy} & 73.12 & 3.44 & 69.10 & 4.89 & 90.40 & 4.06 & 77.54 & 4.13 \\
        \textbf{partly cloudy} & 77.46 & 4.18 & 75.79 & 6.10 & 89.98 & 4.61 & 81.08  & 4.96\\
        \textbf{rain} & 61.88 & 3.94 & 58.00 & 4.32 & 89.79 & 4.65 & 69.89 & 4.30\\
        \textbf{sleet} & 47.66 & 4.17 & 50.00 & 4.75 & 92.57 & 4.51 & 63.41 & 4.48\\
        \textbf{snow} & 29.50 & 1.42 & 33.70 & 4.48 & 89.03 & 6.83 & 50.74 & 4.24\\
        \textbf{tstorms} & 84.50 & 3.88 & 80.75 & 3.32 & 89.09 & 4.58 & 84.78 & 3.93\\
        \textbf{unknown} & 83.36 & 5.76 & 68.27 & 5.29 & 86.17 & 3.84 & 79.27 & 4.96\\
        \textbf{blanks} & 83.09 & 7.08 & 83.84 & 6.38 & 78.93 & 8.94 & 81.95 & 7.47\\       
    
    \midrule
     \textbf{Avg.} & 69.49 & 4.82 & 67.25 & 6.77 & 88.73 & 5.03 & 75.16 & 5.54 \\
        \bottomrule
    \end{tabular*}
    \vspace*{5pt}
    \caption{\small Sky segmentation results break down w.r.t weather on SkyFinder dataset. Numbers are in percentage.}
    \label{tab:weather-results}
      %  \vspace{0pt}
\end{table}
% \subsection{Failure cases}

%-------------------------------------

\subsubsection{Performance analysis for different day times} \label{tod}
We also wanted to see how the trained model performed during different times of day, similar to \cite{C1}. As we use three test splits for reporting our results, we sorted each of them w.r.t hour of the day. Then for each sorted split, we evaluate the respective model on its test set and compute mIOU and MCR. We then calculate average over them (see table \ref{tab:time-of-day-results}). Please note that while evaluating RefineNet-Res101-SkyFinder for each of its respective test set, we used scale=1.0 (i.e. full scale at test time) and report the numbers. We find similar pattern as discussed in \cite{C1} for RefineNet as well i.e., model achieves good performance during day time in terms of MCR, and performance decreases during start(early morning) and end of the day(night time), thus MCR increases. In terms of mIOU, we witness a decrease in performance towards the end of the day, but overall the performance seems consistent. Fig \ref{fig:iou-weather-tod} shows illustration for mIOU during different day hours. For comparing RefineNet with the results in the baseline paper in this aspect, we fetched MCR scores for each split w.r.t hour and report results provided by \cite{C1}. Fig.\ref{fig:time-of-day} shows that although our trained model follows the same pattern, but our pixel-level sky detector performs significantly better than all three methods.

Please see figure \ref{fig:time} for qualitative results when we tested RefineNet-Res101-SkyFinder on sorted test split w.r.t time. Row 1 shows success cases where row2 shows failure cases for different times of day.

\begin{figure}[t]
\begin{center}
\includegraphics[width = 1\linewidth,height=2in]{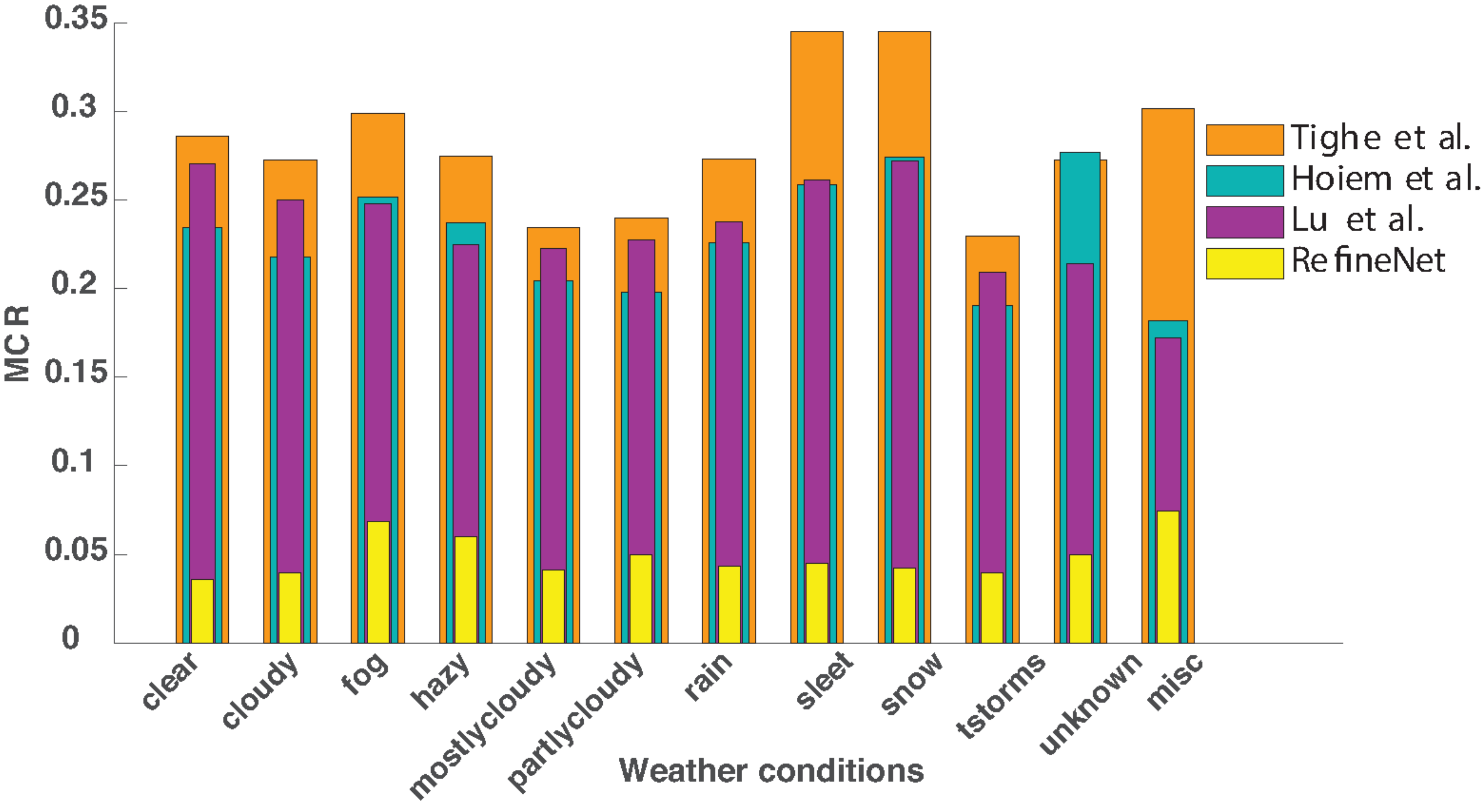}
\end{center}
%\vspace{-15pt}
\caption{\small Illustration of performance analysis on different weather conditions. X-axis shows different weather conditions existing in SkyFinder dataset and y-axis represents mean classification rate(MCR) over all three test splits we used in our experiment.}
\label{fig:weather}
%\vspace{-10pt}
\end{figure}

\subsubsection{Performance analysis for weather conditions} \label{weather}
Like \cite{C1}, we aim to evaluate how well semantic segmentation model when trained as a sky classifier performs for different weather conditions. We split the testset with respect to different weather conditions as provided in metadata of SkyFinder dataset and evaluated RefineNet-Res101-SkyFinder on them. We run the same experiment for all three test splits and then compute the average scores for both mIOU and MCR (See table \ref{tab:weather-results} for quantitative results). Looking at the average mIOU scores, we find that our model struggles for few weather conditions like snow and sleet. On the other hand our model performs better even when weather is cloudy, partly cloudy, or foggy. Interestingly, the model is able to perform really well during thunderstorms both in terms of mIOU and MCR. 

\begin{figure}[t]
\begin{center}
\begin{tabular}{ccccccccc}

\hspace{0pt}Input & \hspace{-5pt} GT & \hspace{-5pt}Pred. &\hspace{-5pt}Input  &\hspace{-5pt} GT & \hspace{-5pt}Pred. & \hspace{-5pt}Input & \hspace{-5pt}GT & \hspace{-5pt}Pred.
\\
\multicolumn{9}{c}{\includegraphics[width=1.0\linewidth]{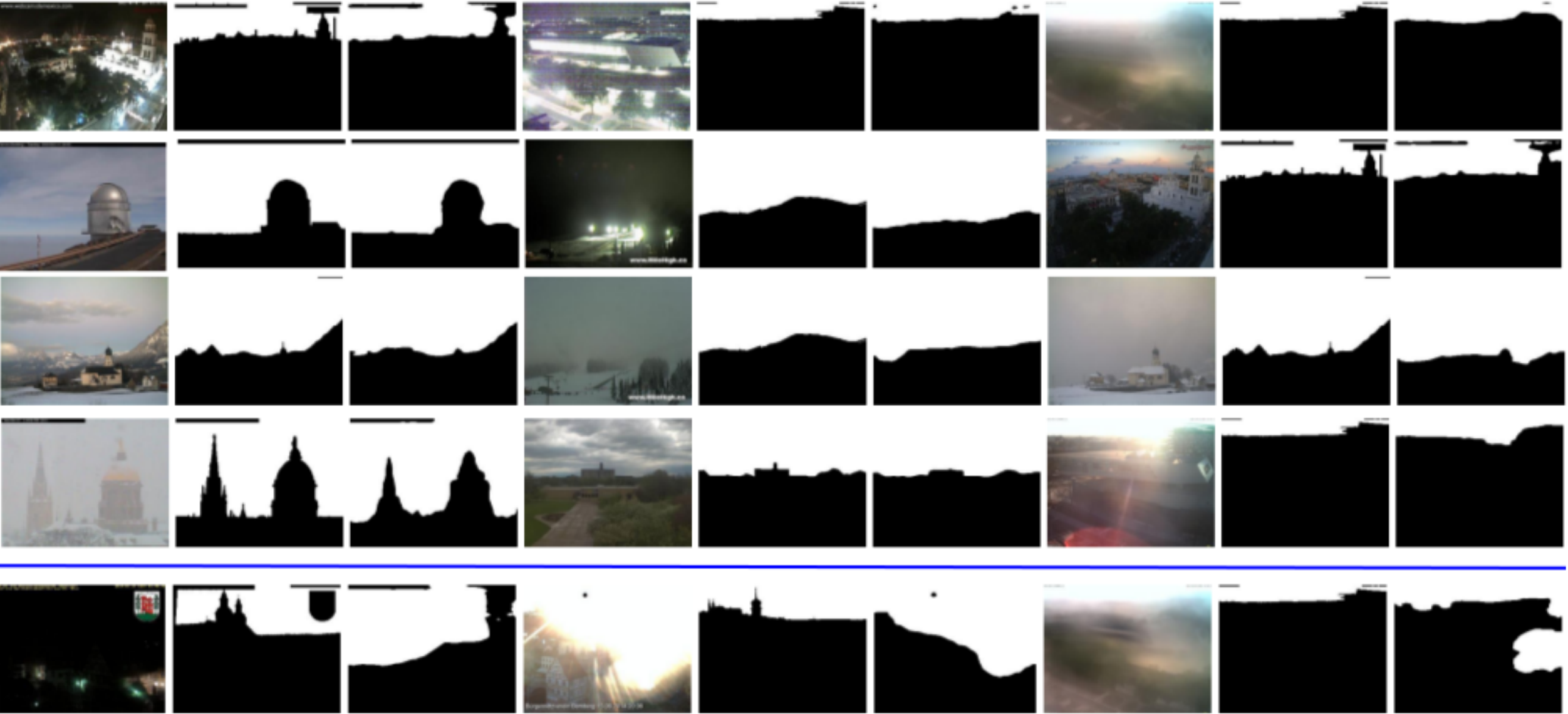}}

\end{tabular}
   
\end{center}
   \caption{\small Qualitative results for performance analysis w.r.t weather on SkyFinder dataset using RefineNet-Res101-SkyFinder. First four rows show success cases(mIOU $>$ 0.8) for each weather type, and last row shows images where model fails(mIOU $<$ 0.5). Success cases in first 4 rows follow the weather order: clear, cloudy, fog, hazy, misc, mostly cloudy, partly cloudy, rain, sleet, snow, tstorms, and unknown respectively. For each weather type, we show input image, ground truth and prediction respectively. Last row shows failure cases where weather type was: tstorms , clear, and fog.}
\label{fig:weather-qual}

\end{figure}

SkyFinder dataset has some images without labels for weather conditions, we evaluated our model on them and report the results. We observe that mIOU is lower than expected for clear sky. This is due to the reason that a big share of clear sky images (1200+) are from camera 10917 where there is no sky visible, thus it lowers the mIOU at test time for splits 1 and 2. Split 3 has been trained on such images and therefore learns to tell when there is no sky in the image (92.61\% mIOU for clear sky).Fig \ref{fig:iou-weather-tod} shows mIOU scores averaged over all three splits for RefineNet-Res101-SkyFinder. Fig \ref{fig:weather} compares MCR score for weather with different methods, and RefineNet significantly outperforms them. For calculating MCR for baseline methods, we use the MCR scores from SkyFinder metadata. This figure reports the average MCR score over all test splits. Please see figure \ref{fig:weather-qual} for qualitative results. % on experiments we performed in our paper. 

%-------figure for weather ----

%-------------------------------------------------------------------------

\subsubsection{Performance analysis w.r.t transient attributes} \label{transient}

Inspired by \cite{C1}, we aimed to investigate how much transient attributes affect the performance of our trained model. Hence, we selected images from one of our test splits for four transient attributes: gloomy, clouds, cold and night. We selected images having high presence of these attributes i.e., thresholded above 0.8 and with low chances of their existence i.e., value is less than 0.2. We then tested RefineNet-Res101-SkyFinder on these subsets of images(2 for each attribute : low and high) and report the performance.

\begin{figure*}[t]
\centering
\includegraphics[width=1\linewidth,height=2.3in]{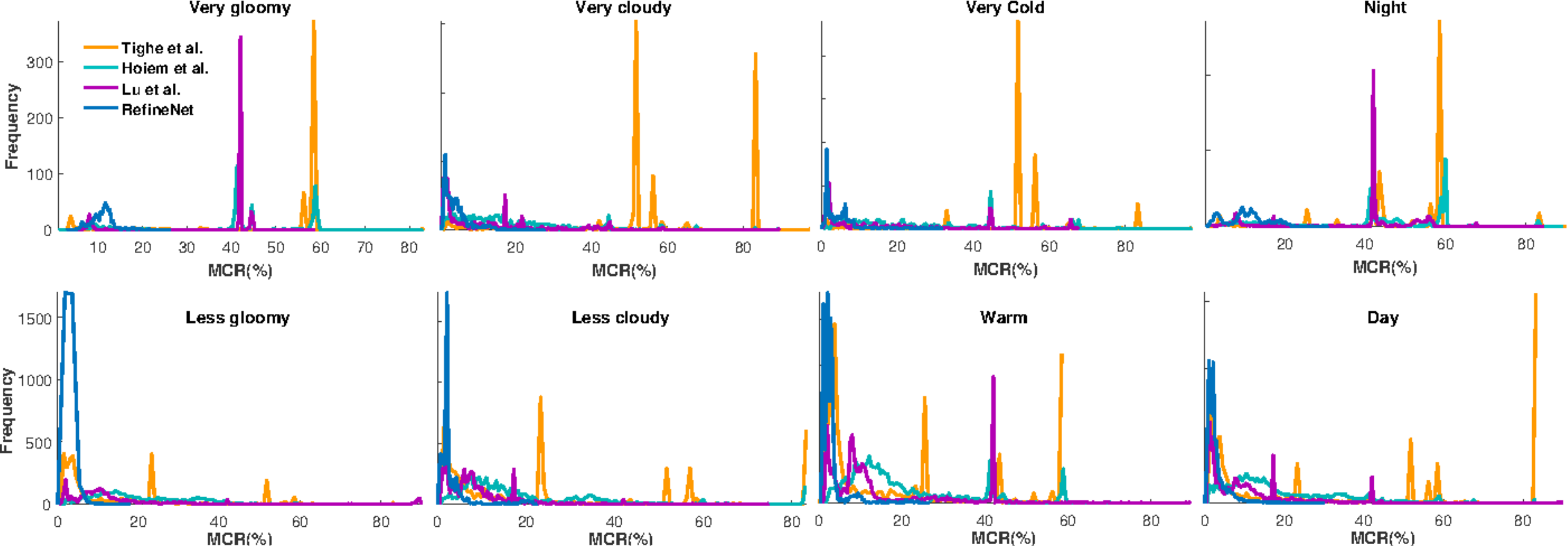}
% \begin{tabular}{cccc}

% \includegraphics[width=0.31\linewidth,height = 1.2in]{images/gloomy0_2.pdf} & \hspace{-30pt}
% \includegraphics[width=0.31\linewidth,height = 1.2in]{images/clouds0_2.pdf} &
% \includegraphics[width=0.31\linewidth,height = 1.2in]{images/cold0_2.pdf} & \hspace{-30pt}
% \includegraphics[width=0.31\linewidth,height = 1.2in]{images/night0_2.pdf} 
% \end{tabular}
%\vspace*{-10pt}
 \caption{\small Comparison of RefineNet-Res101-SkyFinder with baseline methods (Tighe et al., Hoiem et al., and Lu et al.) in terms of MCR given high values($>=$0.8) and low values($<=$0.2) of transient features(gloomy, clouds, cold, and night). Row 1 shows performance comparison when each of four attributes has high value i.e., thresholded above 0.8, whereas row 2 shows results when very low value is observed for each attribute.}
\label{fig:transient_low}
%\vspace{-10pt}
\end{figure*}

\begin{figure*}[t]
\centering
\includegraphics[width=1.0\linewidth,height = 1.4in]{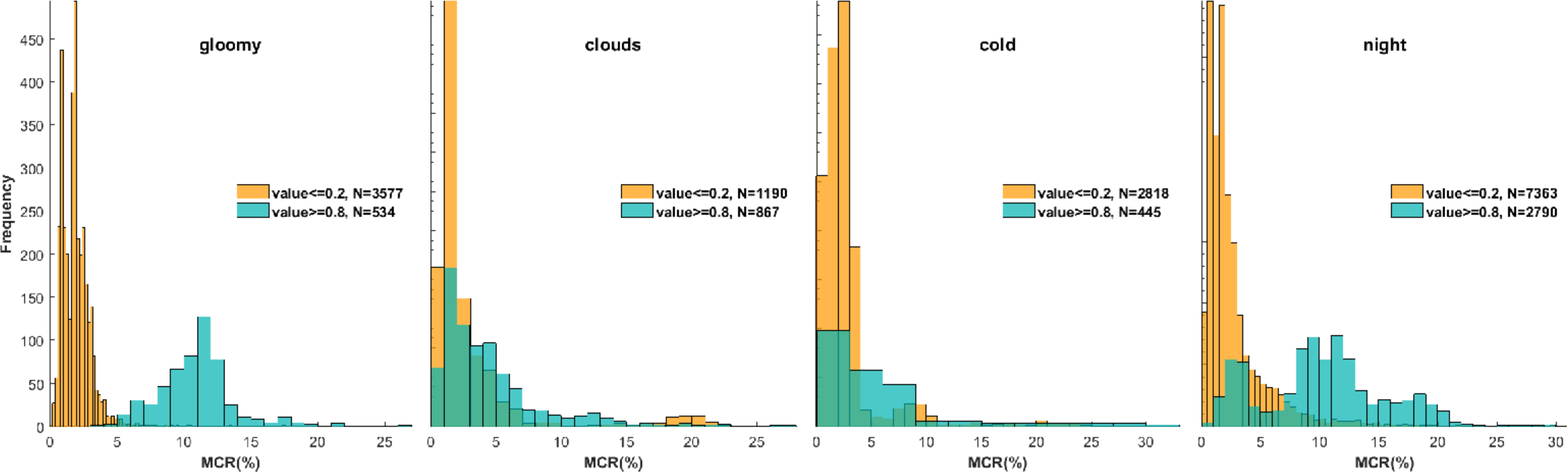}
% \begin{tabular}{cccc}
% \hspace{-15pt}
% \includegraphics[width=0.33\linewidth,height = 1.2in]{images/gloomy.pdf} & \hspace{-60pt}
% \includegraphics[width=0.33\linewidth,height = 1.2in]{images/clouds.pdf} & 
% \hspace{-60pt}
% \includegraphics[width=0.33\linewidth,height = 1.2in]{images/cold.pdf} & \hspace{-60pt}
% \includegraphics[width=0.33\linewidth,height = 1.2in]{images/night.pdf} 
% \end{tabular}
%\vspace*{-10pt}
 \caption{\small Frequency distribution of MCR with RefineNet-Res101-SkyFinder given transient features(gloomy, clouds, cold, and night) using low and high threshold values(0.2 and 0.8, respectively).  }
\label{fig:rn_transient}
%\vspace{-10pt}
\end{figure*}

\begin{figure*}[t]
\begin{center}
\begin{tabular}{cccccc}

\hspace{15pt}Input image & \hspace{40pt} GT & \hspace{45pt}Prediction & \hspace{25pt}Input image & \hspace{40pt}GT & \hspace{20pt}Prediction
\\
\multicolumn{6}{c}{\includegraphics[width=1.0\linewidth]{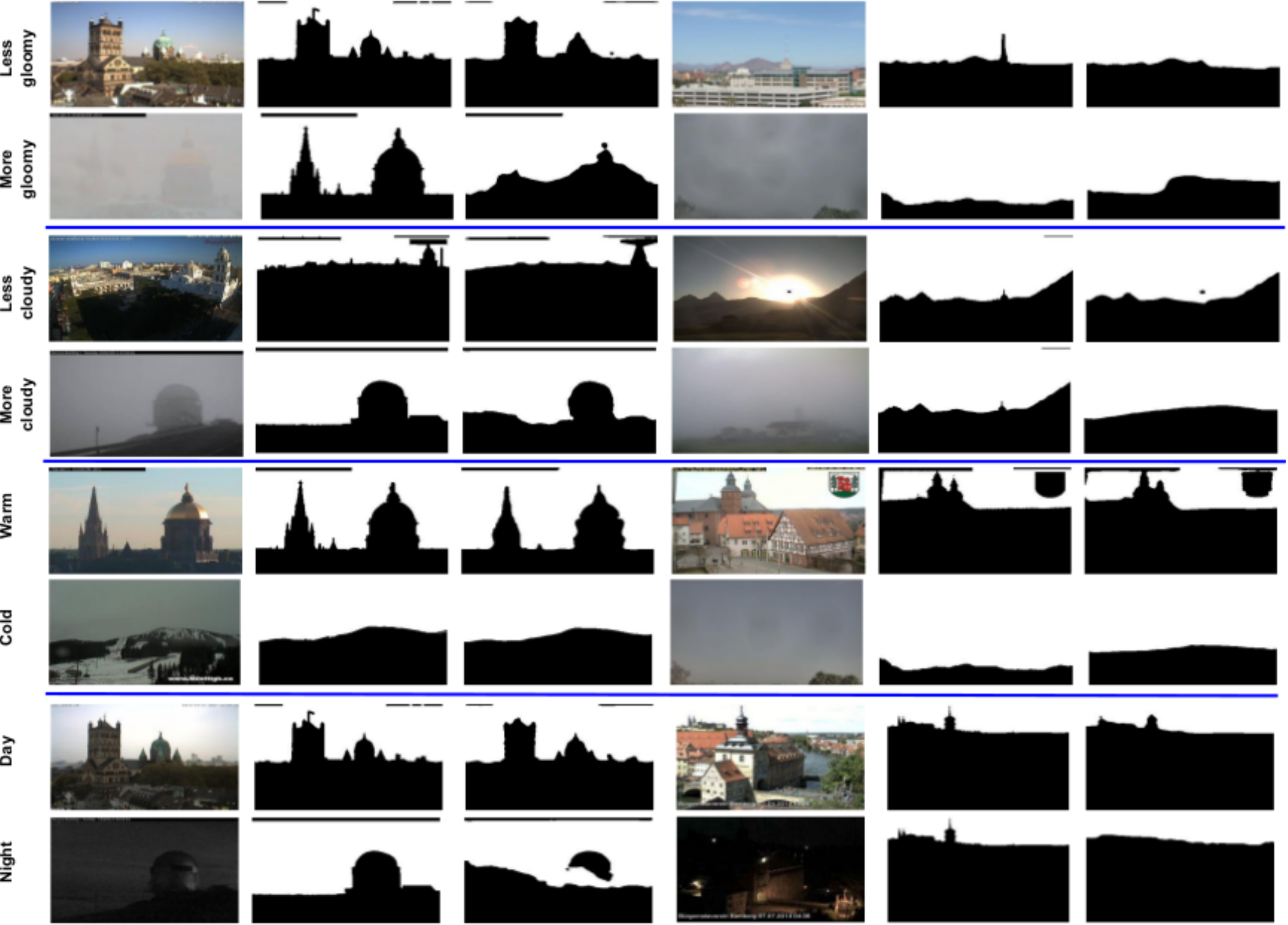}}

\end{tabular}
   
\end{center}
   \caption{Qualitative results for absence(below a threshold i.e. $<=$0.2) and presence(above a threshold i.e. $>=$0.8) of transient attributes (gloomy, clouds, cold, and night) on SkyFinder dataset when using RefineNet-Res101-SkyFinder model. Row 1 shows selected examples for clouds with threshold $<=$0.2, row 2 shows results for clouds using high threshold $>=$0.8, row 3 and 4 show results for gloomy, row 5 and 6 are results for cold, and the last two rows show results for night following the same order as mentioned for first two rows i.e., absence and presence of each transient attribute, respectively.}
\label{fig:transient}

\end{figure*}

\noindent As expected, the model performs well for the images where the sky is not gloomy, cloudy, cold or night time. The model is most robust to clouds and performs well(both in terms of mIOU and MCR) even when clouds are largely present in the image. In terms of MCR, model's performance is worse when we input a highly gloomy image (MCR=10.84) or when it is night time (MCR=10.25). In terms of mIOU, our trained model performs slightly better for gloomy images than night time images (see table \ref{tab:transient}, last row). We also report MCR scores from baseline methods for comparisons and find that Tighe \etal performs poorly among all of the baselines in high presence of the above mentioned transient attributes. Lu \etal seems robust among the baselines but our method significantly outperforms all of them for this experiment(see fig. \ref{fig:transient_low}). Figure \ref{fig:rn_transient} shows the histogram of MCR scores when each of these four transient attributes are high and low in the input images. See figure \ref{fig:transient} for qualitative results of this experiment. 

%---------
% \begin{table}[t]
% \small
% \renewcommand{\arraystretch}{1.0}
%   \centering %\setlength{\tabcolsep}{.56\tabcolsep}   
%     \begin{tabular}{lcccc}
%         \toprule
%                  & \multicolumn{2}{c}{\textbf{$>=$0.8}} & \multicolumn{2}{c}{\textbf{$<=$0.2}}   \\
%                 \cmidrule(lr){2-3}
% \cmidrule(lr){4-5} 
% 		 \textbf{Attribute}        & mIOU  & MCR  & mIOU  & MCR     \\
          
%         \midrule
%     gloomy  & 84.16 & 10.84 & 92.97 & 1.94   \\
%     clouds  & 93.09 & 4.35 & 93.65 & 2.93 \\
%     cold & 90.84 & 5.35 & 92.19 & 2.90 \\
%     night & 83.26 & 10.25 & 94.36  & 2.72 \\
%     \midrule
 
%     \end{tabular}
%    % \vspace*{-10pt}
%     \caption{\small Performance analysis on transient attributes (high thresholded and low thresholded values) using RefineNet-Res101-SkyFinder.}
%     \label{tab:transient}
%     \vspace{0pt}
% \end{table}

%------------------bigger table-----
\begin{table*}[t]
\small
\renewcommand{\arraystretch}{1.0}
  \centering \setlength{\tabcolsep}{.3\tabcolsep}   
    \begin{tabular}{lcccccccccccccccc}
        \toprule
        & \multicolumn{4}{c}{\textbf{gloomy}} & \multicolumn{4}{c}{\textbf{clouds}} & \multicolumn{4}{c}{\textbf{cold}} & \multicolumn{4}{c}{\textbf{night}} \\
        \cmidrule(lr){2-5}  \cmidrule(lr){6-9}  \cmidrule(lr){10-13}  \cmidrule(lr){14-17}
                 & \multicolumn{2}{c}{\textbf{$>=$0.8}} & \multicolumn{2}{c}{\textbf{$<=$0.2}} & \multicolumn{2}{c}{\textbf{$>=$0.8}} & \multicolumn{2}{c}{\textbf{$<=$0.2}} & \multicolumn{2}{c}{\textbf{$>=$0.8}} & \multicolumn{2}{c}{\textbf{$<=$0.2}} & \multicolumn{2}{c}{\textbf{$>=$0.8}} & \multicolumn{2}{c}{\textbf{$<=$0.2}}  \\
                \cmidrule(lr){2-3}
\cmidrule(lr){4-5}  \cmidrule(lr){6-7}  \cmidrule(lr){8-9}  \cmidrule(lr){10-11}  \cmidrule(lr){12-13}  \cmidrule(lr){14-15}  \cmidrule(lr){16-17}  
		 \textbf{Method}        & mIOU  & MCR  & mIOU  & MCR & mIOU  & MCR  & mIOU  & MCR & mIOU  & MCR  & mIOU  & MCR & mIOU  & MCR  & mIOU  & MCR     \\
          
        \midrule
    \textbf{Hoiem \etal}  & - & 40.70 & - & 19.70 & - & 18.47 & - & 18.39 & - & 24.93 & -& 19.23 & - & 41.57 & - & 18.46\\
    \textbf{Lu \etal}  & - & 36.42 & - & 14.88 & - & 12.74 & - & 8.35 & - & 16.44 & - & 14.98 & - & 36.60 & - & 10.03\\
    \textbf{Tighe \etal} & - & 50.53 & - & 13.36 & - & 56.21 & - & 23.44 & - & 50.65 & - & 17.83 & - & 48.14 & - & 31.05 \\
    \textbf{RNet-Res101-SkyFinder} & 84.16&10.84&92.97&1.94 & 93.09&4.35&93.65&2.93 & 90.84&5.35&92.19&2.90 & 83.26&10.25&94.36&2.72 \\
    \midrule
 
    \end{tabular}
   % \vspace*{-10pt}
    \caption{\small Performance analysis on transient attributes (high thresholded and low thresholded values).}
    \label{tab:transient}
    \vspace{0pt}
\end{table*}

%----------------------------------------------------------------------
%------------------------------------------------------------------------------
%--------------------------
% \begin{figure*}[t]
% \centering
% \begin{tabular}{cc}

% \includegraphics[width=0.51\linewidth,height = 2in]{images/gloomy0_8.pdf} & \hspace{-30pt}
% \includegraphics[width=0.51\linewidth,height = 2in]{images/clouds0_8.pdf} 
% \\
% \includegraphics[width=0.51\linewidth,height = 2in]{images/cold0_8.pdf} & \hspace{-30pt}
% \includegraphics[width=0.51\linewidth,height = 2in]{images/night0_8.pdf} 
% \end{tabular}
% %\vspace*{-10pt}
%  \caption{\small Comparison of RefineNet-Res101-SkyFinder with baseline methods (Tighe et al., Hoiem et al.,and Lu et al.) in terms of MCR given high values($>=$0.8) of transient features(gloomy, clouds, cold, and night).}
% \label{fig:transient_high}
% %\vspace{-10pt}
% \end{figure*}

%-----------------------------------------------------------------------
%--------------------------------------------------
\subsubsection{Performance analysis w.r.t noisy images} \label{noise}
\begin{figure}[t]
\begin{center}
\begin{tabular}{cccccc}

\hspace{8pt} \scriptsize Image & \hspace{-2pt} \scriptsize Motion blur & \hspace{-5pt} \scriptsize Gaussian & \hspace{1pt} \scriptsize Poisson & \hspace{-5pt} \scriptsize Salt \& Pepper & \hspace{-8pt} \scriptsize Speckle
\\ 
\multicolumn{6}{c}{\includegraphics[width=1.0\linewidth]{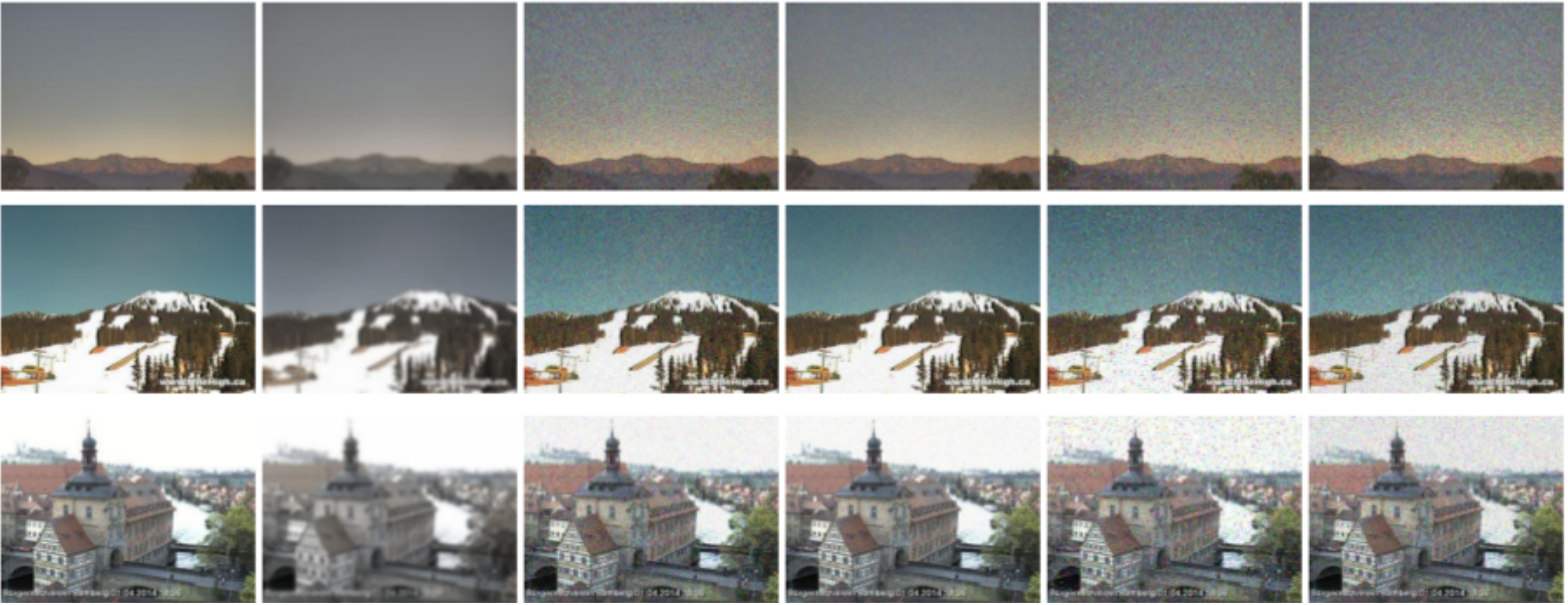}}

\end{tabular}
   
\end{center}
\vspace{-10pt}
   \caption{\small Some example images for different types of noise added to the selected subset of images from SkyFinder dataset.}
\label{fig:noise}

\end{figure}

We are also interested in investigating the impact of noise on the task of Sky segmentation. We conducted this experiment since the image can be noisy due to various reasons in original settings as well. For this purpose, we selected 50 images for each camera from one of the test splits (total 650 images) and added different types of noise to them. We added motion blur, Gaussian noise, Poisson noise, salt \& pepper, and speckle noise to our subset of images. We first evaluated performance without adding any noise and then compared our results with noisy images(see table \ref{tab:noise}). We find that our trained model is robust to Poisson noise, and highly sensitive to salt \& pepper noise. Motion blur also affects mIOU (a drop of approx. 3.5\%). Similarly, Gaussian noise and speckle noise also hurt performance (both in terms of mIOU and MCR). See figure \ref{fig:noise} for sample images after adding different types of noise to input images.

\begin{table}[t]
\footnotesize
\renewcommand{\arraystretch}{.9}
  \centering \setlength{\tabcolsep}{.17\tabcolsep}   
    \begin{tabular}{lcccccc}
        \toprule               
		          & Original  & Motion Blur  & Gaussian  & Poisson & Salt \& Pepper & Speckle \\
         \cmidrule(lr){2-2} \cmidrule(lr){3-3} \cmidrule(lr){4-4}\cmidrule(lr){5-5}\cmidrule(lr){6-6}\cmidrule(lr){7-7}
        
    mIOU  & 89.92 & 86.31 & 85.26 & 88.67 & 81.77 & 84.70\\
    MCR  & 5.08 & 6.65 & 7.60 & 5.73 & 9.32 & 7.91\\
  
   \bottomrule
 
    \end{tabular}
    \vspace*{2pt}
    \caption{\small Performance analysis on SkyFinder dataset when input is a noisy image. Column 1 shows results on images without any noise, whereas rest of the columns show results when we added motion blur, Gaussian noise, Poisson noise, salt \& pepper noise, and speckle noise to the images respectively.}
    \label{tab:noise}
    \vspace{0pt}
\end{table}

%-----------------------------------------------------------------------

%------------------------------------------

\subsection{Sun-Sky dataset}
%---------------------------------------------------------------
The SUN-Sky dataset is a subset of images from SUN dataset \cite{xiao2010sun} of outdoor scenes having sky present in them. Using a similar approach as we used for training on SkyFinder dataset, we first evaluate RefineNet-Res101-Cityscapes model on SUN-Sky dataset, then fine-tune the same model on this dataset referred as RefineNet-Res101-SUNdb-FT, and lastly, we trained RefineNet-Res101- on SUN-Sky dataset from the scratch (referred to as RefineNet-Res101-SUNdb). We randomly split the images into 60\%-20\%-20\% ratio for training, validation and testing respectively. When evaluated these three models on our test set, we find that off-the-shelf model performs better(mIOU = 61.69, MCR = 8.4) on this dataset as compared to its performance on SkyFinder dataset (mIOU = 48.5, MCR = 18.93), which maintains that SkyFinder is more challenging in nature as compared to SUN-Sky dataset. Interestingly, the fine-tuned model on this dataset outperforms the model when trained solely on SUN-Sky dataset with a slight margin (both in terms of mIOU and MCR) which is mainly because the SUN-Sky dataset is not big enough in size. Overall, we find that both fine-tuned and trained models on this dataset performs reasonably well as compared to off-the-shelf RefineNet-Res101-Cityscapes model (see table \ref{tab:SUNData}).

\subsection{Cross datasets evaluation} \label{cross_dataset}
To see generalization power of our models, we evaluated them cross datasets i.e., models trained on SkyFinder were evaluated on SUN-Sky dataset and vice versa. Table \ref{tab:diff-datasets-results} shows results for both fine-tuned and trained model on each dataset and across other dataset. Looking at the results, we find that SkyFinder, due to its large size, gives better performance when we trained the model on it from the scratch. Interestingly, our fine-tuned models perform better than trained from scratch when evaluated across datasets. In terms of MCR, fine-tuned model on SkyFinder generalizes better than the model fine-tuned on SUN-Sky dataset(6.67 vs. 11.74). Please note, for SkyFinder dataset, results shown from all models are averaged over all test splits.
\begin{table}[t]
\small
  \centering \setlength{\tabcolsep}{.3\tabcolsep} 
  \renewcommand{\arraystretch}{.9}
  
    \begin{tabular}{lcccc}
        \toprule
                \multicolumn{1}{l}{\textbf{Datasets}} & \multicolumn{2}{c}{\textbf{SkyFinder}} & \multicolumn{2}{c}{\textbf{SUN-Sky}}   \\
                \cmidrule(lr){2-3}
\cmidrule(lr){4-5} 
          & mIOU(\%)  & MCR(\%)  & mIOU(\%)  & MCR(\%) \\         
        \midrule
    \small RNet-SkyFinder-FT & 79.00& 6.69 & \textcolor{blue}{71.70} & \textcolor{blue}{6.67} \\  
   
   \small RNet-SkyFinder  & \textbf{83.00} & \textbf{5.89} & 71.57 & 7.24 \\
   
  \small  RNet-SUNdb-FT & \textcolor{blue}{71.49} & \textcolor{blue}{11.74} & \textbf{83.10} & \textbf{3.70} \\
   \small RNet-SUNdb & 70.42 & 12.37 & 82.36 & 4.17 \\
        \bottomrule
    \end{tabular}
    \vspace*{5pt}
    \label{tab:SUN-Sky}
    \caption{\small Sky segmentation results on 2 datasets broken down and trained on one sub dataset and tested on others. Numbers in bold text show the best results on particular dataset, whereas, blue font shows best results for that dataset during cross dataset evaluation. Numbers are in percentage.}
    \label{tab:diff-datasets-results}
        \vspace{-10pt}
\end{table}

%---------------------------------------------------------------------
\section{Discussion and Conclusion} \label{conclusion}

We will first direct our focus to the original results of RefineNet's res101 model trained on Cityscapes in Table 1. Both results incorporate the pretrained model's incapability to properly access non-ideal images, particularly night images as addressed earlier. Upon finetuning on the model using the SkyFinder dataset, we note a drastic improvement in the mIOU and just over 10\% decrease in the MCR. This is clearly as a result of including the non-ideal images. Upon training it on the SkyFinder dataset, we see a less drastic improvement in one split, but improvement nonetheless. 

Despite the averages in Table 1, focusing on a singular camera in one of the data splits, we can see a drastic improvement. Camera 858, consists of 200 images and was not seen by the finetuned or ImageNet initialized model during training. The mIOU from the model trained on Cityscapes was 69.21\%, and had an MCR of 18.16\%. However, after finetuning the model, the mIOU and MCR results respectively quickly jumped up to 91.41\% and 4.69\%. Finally, after training on only 28 cameras of the data (again, not including camera 858) and initializing from ImageNet, the results improved slightly. The mIOU increased to 95.76\% and the MCR dropped a little more to 2.30\%. Other cameras show similar rates of improvement. There are some cameras however that prove difficult to segment for all methods and show a smaller rate of improvement.

The importance of these results lies in the ability to use these models in real-world applications. Using the original pre-trained model would result in poor quality segmentation outside of the ideal circumstances. Off-the-shelf methods must be modified in order to be used most effectively in the real world.

Understanding the impacts of these results, we also incorporate the results from Mihail et al's findings for their baseline methods. While their own model's results which reported an MCR of 12.96\% across their own testing split has not been used to report the performance in terms of MCR on our test splits. In spite of this, we can still further prove the idea that existing models are still effective--so long as they have been modified to suit the task's needs. 

We also demonstrated that, overall, even state-of-the-art models struggle with challenging conditions like night time, variation in weather and other transient attributes. Although, our trained models still perform much better than prior methods in terms of MCR.

%\section{Future Work}

Following this work, we intend to look at other scene parsing models to evaluate their off-the-shelf methods, finetune them, and possibly train them on the SkyFinder dataset in order to compare the results to the above. We also plan to develop our own end-to-end sky segmentation model to also use for comparison. To possibly improve general results, cleaning of the dataset may need to occur, such as removing timestamps. Other future directions include using sky segmentation for applications such as weather classification and weather forecasting. Code, our trained models and data will be made available for further exploration of this area.

{\small
\bibliographystyle{ieee}
\bibliography{egbib}
}

\end{document}